\documentclass{article}

\usepackage[final,nonatbib]{neurips_2023}

\usepackage{microtype}      
\usepackage{graphicx}
\usepackage{tikz}
\usepackage{xfp}
\usepackage{url}            
\usepackage{booktabs}       
\usepackage{amsfonts}       
\usepackage{nicefrac}       
\usepackage{xcolor}         
\usepackage{amsmath}
\usepackage{amssymb} 
\usepackage{amsthm}         
\usepackage{multirow}
\usepackage{array}          
\usepackage[pagebackref,hidelinks]{hyperref}

\theoremstyle{plain}
\newtheorem{theorem}{Theorem}[section]
\newtheorem{proposition}[theorem]{Proposition}

\theoremstyle{definition}

\theoremstyle{remark}

 \renewcommand{\hat}{\widehat}

 \newcommand{\defn}{\triangleq}

 \newcommand{\uvec}[1]{\ensuremath{\underline{\boldsymbol{#1}}}}
 \newcommand{\ovec}[1]{\boldsymbol{\Bar{#1}}}
 \newcommand{\hvec}[1]{\ensuremath{\Hat{\boldsymbol{#1}}}}
 \renewcommand{\vec}[1]{\ensuremath{\boldsymbol{#1}}}
 \newcommand{\huvec}[1]{\ensuremath{\Hat{\underline{\boldsymbol{#1}}}}}

 \newcommand{\mc}[1]{\ensuremath{\mathcal{#1}}}

 \newcommand{\Real}{{\mathbb{R}}}
 \newcommand{\Complex}{{\mathbb{C}}}

 \newcommand{\Ell}{\mathcal{L}}

 \newcommand{\tran}{^{\top}}
 \newcommand{\herm}{^{\textsf{H}}}
 \newcommand*\deriv{\mathop{}\!\mathrm{d}}

 \DeclareMathOperator{\real}{Re}

 \DeclareMathOperator{\E}{E}
 \DeclareMathOperator{\Cov}{Cov}

 \DeclareMathOperator{\tr}{tr}

 \DeclareMathOperator{\erf}{erf}

 \renewcommand{\eqref}[1]{(\ref{eq:#1})}
 
 \newcommand{\Figref}[1]{Figure~\ref{fig:#1}}
 \newcommand{\figref}[1]{Fig.~\ref{fig:#1}}
 \newcommand{\tabref}[1]{Table~\ref{tab:#1}}
 \newcommand{\secref}[1]{Section~\ref{sec:#1}}
 \newcommand{\appref}[1]{Appendix~\ref{app:#1}}

 \newcommand{\propref}[1]{Proposition~\ref{prop:#1}}

 \newcommand{\mmse}{_{\mathsf{mmse}}}
 \newcommand{\pa}{p_{\mathsf{a}}}
 \newcommand{\pb}{p_{\mathsf{b}}}
 \newcommand{\pab}{p_{\mathsf{a,b}}}
 
 \newcommand{\py}{p_{\mathsf{y}}}

 \newcommand{\pxgy}{p_{\mathsf{x}|\mathsf{y}}}
 \newcommand{\pz}{p_{\mathsf{z}}}
 
 \newcommand{\pxhgy}{p_{\hat{\mathsf{x}}|\mathsf{y}}}
 \newcommand{\Eab}{\E_{\mathsf{a,b}}}
 \newcommand{\Ey}{\E_{\mathsf{y}}}
 \newcommand{\Exgy}{\E_{\mathsf{x}|\mathsf{y}}}
 \newcommand{\Exy}{\E_{\mathsf{x,y}}}
 \newcommand{\Exhgy}{\E_{\hat{\mathsf{x}}|\mathsf{y}}}
 \newcommand{\Egy}{\E_{\hat{\mathsf{x}},y}}
 \newcommand{\Ez}{\E_{\mathsf{z}}}
 
 \newcommand{\Exzy}{\E_{\mathsf{x,z,y}}}
 \newcommand{\Exzzy}{\E_{\mathsf{x,z_1,z_2,y}}} 
 \newcommand{\ExzPy}{\E_{\mathsf{x,z_1,...,z_P,y}}}

 \newcommand{\Ezigy}{\E_{\mathsf{z_i}|\mathsf{y}}}
 \newcommand{\EzPgy}{\E_{\mathsf{z_1,...,z_P}|\mathsf{y}}}
 \newcommand{\EzPy}{\E_{\mathsf{z_1,...,z_P,y}}}
 \newcommand{\ExzPgy}{\E_{\mathsf{x,z_1,...,z_P}|\mathsf{y}}}
 \newcommand{\ExxP}{\E_{X,\hat{X}_1,...\hat{X}_P}}
 \newcommand{\ExP}{\E_{\hat{X}_1,...\hat{X}_P}}
 \newcommand{\Covzigy}{\Cov_{\mathsf{z_i}|\mathsf{y}}}
 \newcommand{\Covxgy}{\Cov_{\mathsf{x}|\mathsf{y}}}
 
 \newcommand{\raw}{^{\mathsf{raw}}}
 \newcommand{\batch}{_{\mathsf{batch}}}
 
 \newcommand{\train}{_{\mathsf{train}}}
 
 \newcommand{\val}{_{\mathsf{val}}}
 \DeclareMathOperator{\CWD}{CWD}
 \newcommand{\Dadler}{D^{\mathsf{adler}}}
 \newcommand{\Ladler}{\mc{L}^{\mathsf{adler}}_{\mathsf{adv}}}
 \newcommand{\Ladv}{\mc{L}_{\mathsf{adv}}}
 \newcommand{\Lgp}{\mc{L}_{\mathsf{gp}}}
 \newcommand{\Ldrift}{\mc{L}_{\mathsf{drift}}}
 \newcommand{\Ltwo}{\mc{L}_2}
 \newcommand{\LtwoP}{\mc{L}_{2,P}}
 \newcommand{\LtwoPt}{\mc{L}_{2,P\train}}
 \newcommand{\LvarP}{\mc{L}_{\mathsf{var},P}}
 \newcommand{\LvarPt}{\mc{L}_{\mathsf{var},P\train}}
 
 \newcommand{\LtwovarPt}{\mc{L}_{2,\mathsf{var},P\train}}
 \newcommand{\LoneP}{\mc{L}_{1,P}}
 \newcommand{\LonePt}{\mc{L}_{1,P\train}}
 \newcommand{\LstdP}{\mc{L}_{\mathsf{SD},P}}
 \newcommand{\LstdPt}{\mc{L}_{\mathsf{SD},P\train}}
 
 \newcommand{\LonestdPt}{\mc{L}_{1,\mathsf{SD},P\train}}
 \newcommand{\betaadv}{\beta_{\mathsf{adv}}}
 
 \newcommand{\betastd}{\beta_{\mathsf{SD}}}
 \newcommand{\betastdG}{\beta_{\mathsf{SD}}^\mc{N}}
 \newcommand{\mustd}{\mu_{\mathsf{SD}}}
 \newcommand{\avgP}{_{\scriptscriptstyle (P)}}
 \newcommand{\avgone}{_{\scriptscriptstyle (1)}}
 \newcommand{\avgPt}{_{\scriptscriptstyle (P\train)}}

 \DeclareMathOperator{\CFID}{CFID}
 \newcommand{\pxugyu}{p_{\underline{\mathsf{x}}|\underline{\mathsf{y}}}}
 \newcommand{\pxuhgyu}{p_{\hat{\underline{\mathsf{x}}}|\underline{\mathsf{y}}}}
 \newcommand{\muxu}{\vec{\mu}_{\underline{\mathsf{x}}}}
 \newcommand{\muxhu}{\vec{\mu}_{\underline{\hat{\mathsf{x}}}}}
 \newcommand{\muyu}{\vec{\mu}_{\underline{\mathsf{y}}}}
 \newcommand{\muxugyu}{\vec{\mu}_{\underline{\mathsf{x}}|\underline{\mathsf{y}}}}
 \newcommand{\muxhugyu}{\vec{\mu}_{\underline{\hat{\mathsf{x}}}|\underline{\mathsf{y}}}}
 \newcommand{\Sigxugyu}{\vec{\Sigma}_{\underline{\mathsf{x}}\underline{\mathsf{x}}|\underline{\mathsf{y}}}}
 \newcommand{\Sigxuhgyu}{\vec{\Sigma}_{\underline{\hat{\mathsf{x}}}\underline{\hat{\mathsf{x}}}|\underline{\mathsf{y}}}}
 \newcommand{\Sigxuxu}{\vec{\Sigma}_{\underline{\mathsf{x}}\underline{\mathsf{x}}}}
 \newcommand{\Sigxhuxhu}{\vec{\Sigma}_{\underline{\hat{\mathsf{x}}}\underline{\hat{\mathsf{x}}}}}
 \newcommand{\Sigyuyu}{\vec{\Sigma}_{\underline{\mathsf{y}}\underline{\mathsf{y}}}}
 \newcommand{\Sigxuyu}{\vec{\Sigma}_{\underline{\mathsf{x}}\underline{\mathsf{y}}}}
 \newcommand{\Sigxhuyu}{\vec{\Sigma}_{\underline{\hat{\mathsf{x}}}\underline{\mathsf{y}}}}

 \newcommand{\zm}{_{\mathsf{zm}}}
 \newcommand{\mean}{_{\mathsf{mean}}}
 \newcommand{\cova}{_{\mathsf{cov}}}

\title{A Regularized Conditional GAN for Posterior Sampling in Image Recovery Problems}

\author{%
  Matthew C. Bendel \\
  Dept. ECE\\
  The Ohio State University\\
  Columbus, OH 43210 \\
  \texttt{bendel.8@osu.edu} \\
  \And
  Rizwan Ahmad \\
  Dept. BME\\
  The Ohio State University\\
  Columbus, OH 43210 \\
  \texttt{ahmad.46@osu.edu} \\
  \And
  Philip Schniter \\
  Dept. ECE\\
  The Ohio State University\\
  Columbus, OH 43201 \\
  \texttt{schniter.1@osu.edu} \\
}

\begin{document}

\maketitle

\begin{abstract}
In image recovery problems, one seeks to infer an image from distorted, incomplete, and/or noise-corrupted measurements.
Such problems arise in magnetic resonance imaging (MRI), computed tomography, deblurring, super-resolution, inpainting, phase retrieval, image-to-image translation, and other applications.
Given a training set of signal/measurement pairs, we seek to do more than just produce one good image estimate.  
Rather, we aim to rapidly and accurately sample from the posterior distribution.  To do this,
we propose a regularized conditional Wasserstein GAN that generates dozens of high-quality posterior samples per second.  
Our regularization comprises an $\ell_1$ penalty and an adaptively weighted standard-deviation reward.
Using quantitative evaluation metrics like conditional Fr\'{e}chet inception distance, we demonstrate that our method produces state-of-the-art posterior samples in both multicoil MRI and large-scale inpainting applications.
The code for our model can be found here: \url{https://github.com/matt-bendel/rcGAN}.
\end{abstract}

\section{Introduction} \label{sec:intro}
We consider image recovery, where one observes measurements $\vec{y}=\mc{M}(\vec{x})$ of the true image $\vec{x}$ that may be masked, distorted, and/or corrupted $\vec{x}$ with noise, and the goal is to infer $\vec{x}$ from $\vec{y}$. 
This includes linear inverse problems arising in, e.g., deblurring, super-resolution, inpainting, colorization, computed tomography (CT), and magnetic resonance imaging (MRI), where $\vec{y}=\vec{Ax}+\vec{w}$ with known linear operator $\vec{A}$ and noise $\vec{w}$.
But it also includes non-linear inverse problems like those arising in phase-retrieval and dequantization, as well as image-to-image translation problems.
In all cases, it is impossible to perfectly infer $\vec{x}$ from $\vec{y}$.

Image recovery is often posed as finding the single ``best'' recovery $\hvec{x}$, which is known as a \emph{point estimate} of $\vec{x}$ \cite{Lehmann:Book:06}.
But point estimation is problematic due to the \emph{perception-distortion tradeoff} \cite{Blau:CVPR:18}, which establishes a fundamental tradeoff between distortion (defined as some distance between $\hvec{x}$ and $\vec{x}$) and perceptual quality (defined as some distance between $\hvec{x}$ and the set of clean images).
For example, the minimum mean-squared error (MMSE) recovery $\hvec{x}\mmse$ is optimal in terms of $\ell_2$ distortion, but can be unrealistically smooth.
Although one could instead compute an approximation of the maximum a posteriori (MAP) estimate \cite{Sonderby:ICLR:17} or minimize some combination of perceptual and distortion losses,
it's unclear which combination would be most appropriate.

Another major limitation with point estimation is that it's unclear how certain one can be about the recovered $\hvec{x}$.
For example, with deep-learning-based recovery, it's possible to hallucinate a nice-looking $\hvec{x}$, but is it correct?
Quantifying the uncertainty in \hvec{x} is especially important in medical applications such as MRI, where a diagnosis must be made based on the measurements \vec{y}. 
Rather than simply reporting our best guess of whether a pathology is present or absent based on \hvec{x}, we might want to report the probability that the pathology is present (given all available data).

Yet another problem with point estimation is that the estimated \hvec{x} could pose issues with fairness \cite{Mehrabi:CS:21}. 
For example, say we are inpainting a face within an image. 
With a racially heterogeneous prior distribution, $\hvec{x}\mmse$ (being the posterior mean) will be biased towards the most predominant race.
The same will be true of many other point estimates $\hvec{x}$.

To address the aforementioned limitations of point estimation, we focus on \emph{generating samples from the posterior distribution} $\pxgy(\cdot|\vec{y})$, which represents the complete state-of-knowledge about $\vec{x}$ given the measurements $\vec{y}$.
The posterior correctly fuses prior knowledge with measurement knowledge, thereby alleviating any concerns about fairness (assuming the data used to represent the prior is fairly chosen \cite{Mehrabi:CS:21}).
Furthermore, the posterior directly facilitates uncertainty quantification via, e.g., pixel-wise standard deviations or pathology detection probabilities (see \appref{detection}).  
Also, if it was important to report a single ``good'' recovery, then posterior sampling leads to an easy navigation of the perception-distortion tradeoff.
For example, averaging $P\geq 1$ posterior samples gives a close approximation to the less-distorted-but-oversmooth $\hvec{x}\mmse$ with large $P$ and sharp-but-more-distorted $\hvec{x}$ with small $P$.
Additionally, posterior sampling unlocks other important capabilities such as adaptive acquisition \cite{Sanchez:NIPSW:20} and counterfactual diagnosis \cite{Chang:ICLR:19}.

Concretely, given a training dataset of image/measurement pairs $\{(\vec{x}_t,\vec{y}_t)\}_{t=1}^T$, 
our goal is to learn a generating function $G_{\vec{\theta}}$ that, for a new $\vec{y}$, maps random code vectors $\vec{z}\sim\mc{N}(\vec{0},\vec{I})$ to posterior samples $\hvec{x}=G_{\vec{\theta}}(\vec{z},\vec{y})\sim\pxgy(\cdot|\vec{y})$.
%
%
There exist several well-known approaches to this task, with recent literature focusing on
conditional generative adversarial networks (cGANs) 
\cite{Isola:CVPR:17,Adler:18,Zhao:ICLR:21,Zhao:MIA:18},
conditional variational autoencoders (cVAEs) 
\cite{Edupuganti:TMI:20,Tonolini:JMLR:20,Sohn:NIPS:15},
conditional normalizing flows (cNFs) 
\cite{Ardizzone:19,Winkler:19,Sun:AAAI:21},
and score/diffusion/Langevin-based generative models 
\cite{Welling:ICML:11,Song:NIPS:20,Jalal:NIPS:21,Song:ICLR:21,Song:ICLR:22}.  
Despite it being a longstanding problem, posterior image sampling remains challenging.
Although score/diffusion/Langevin approaches have dominated the recent literature due to advances in accuracy and diversity, their sample-generation speeds remain orders-of-magnitude behind those of cGANs, cVAEs, and cNFs.

We choose to focus on cGANs, which are typically regarded as generating samples of high quality but low diversity.
Our proposed cGAN tackles the lack-of-diversity issue using a novel regularization that consists of supervised-$\ell_1$ loss plus an adaptively weighted standard-deviation (SD) reward. 
This is not a heuristic choice; we prove that our regularization enforces consistency with the true posterior mean and covariance under certain conditions.

Experimentally, we demonstrate our regularized cGAN on accelerated MRI and large-scale face completion/inpainting. 
We consider these applications for three main reasons.
First, uncertainty quantification in MRI, and fairness in face-generation, are both of paramount importance. 
Second, posterior-sampling has been well studied for both applications, and fine-tuned cGANs \cite{Zhao:ICLR:21} and score/Langevin-based approaches \cite{Jalal:NIPS:21,Song:ICLR:21} are readily available.
Third, the linear operator ``$\vec{A}$'' manifests very differently in these two applications,\footnote{In MRI, the forward operator acts locally in the frequency domain but globally in the pixel domain, while in inpainting, the operator acts locally in the pixel domain but globally in the frequency domain.}
which illustrates the versatility of our approach.
To quantify performance, we focus on conditional Fr\'{e}chet inception distance (CFID) \cite{Soloveitchik:21}, 
which is a principled way to quantify the difference between two high-dimensional posterior distributions,
although we also report other metrics.
Our results show the proposed regularized cGAN (rcGAN) outperforming existing cGANs \cite{Adler:18,Ohayon:ICCVW:21,Zhao:ICLR:21} and the score/diffusion/Langevin approaches from \cite{Jalal:NIPS:21} and \cite{Song:ICLR:21} in all tested metrics, while generating samples $\sim 10^4$ times faster than \cite{Jalal:NIPS:21,Song:ICLR:21}.

%

\section{Problem formulation and background}

We build on the Wasserstein cGAN framework from \cite{Adler:18}. 
The goal is to design a generator network $G_{\vec{\theta}}:\mc{Z}\times\mc{Y}\rightarrow\mc{X}$ such that, for fixed $\vec{y}$, the random variable $\hvec{x}=G_{\vec{\theta}}(\vec{z},\vec{y})$ induced by $\vec{z}\sim\pz$ has a distribution that best matches the posterior $\pxgy(\cdot|\vec{y})$ in Wasserstein-1 distance.
Here,  $\mc{X}$, $\mc{Y}$, and $\mc{Z}$ denote the sets of $\vec{x}$, $\vec{y}$, and $\vec{z}$, respectively, and $\vec{z}$ is drawn independently of $\vec{y}$. 

The Wasserstein-1 distance can be expressed as 
\begin{align}
W_1(\pxgy(\cdot,\vec{y}),\pxhgy(\cdot,\vec{y})) = \sup_{D\in L_1} \Exgy\{D(\vec{x},\vec{y})\} - \Exhgy\{D(\hvec{x},\vec{y})\}
\label{eq:W1},
\end{align}
where $L_1$ denotes functions that are 1-Lipschitz with respect to their first argument and $D:\mc{X}\times\mc{Y}\rightarrow \Real$ is a ``critic'' or ``discriminator'' that tries to distinguish between true $\vec{x}$ and generated $\hvec{x}$ given $\vec{y}$.
Since we want the method to work for typical values of $\vec{y}$, we define a loss by taking an expectation of \eqref{W1} over $\vec{y}\sim\py$.
Since the expectation commutes with the supremum in \eqref{W1},
we have \cite{Adler:18}
\begin{align}
\Ey\{W_1(\pxgy(\cdot,\vec{y}),\pxhgy(\cdot,\vec{y}))\} &= \sup_{D\in L_1} \Exy\{D(\vec{x},\vec{y})\}  - \Egy\{D(\hvec{x},\vec{y})\} \\
&= \sup_{D\in L_1} \Exzy\{D(\vec{x},\vec{y})  - D(G_{\vec{\theta}}(\vec{z},\vec{y}),\vec{y})\}
\label{eq:EW1} .
\end{align}
In practice, $D$ is implemented by a neural network $D_{\vec{\phi}}$ with parameters $\vec{\phi}$, and $(\vec{\theta},\vec{\phi})$ are trained by alternately minimizing
\begin{align}
\Ladv(\vec{\theta},\vec{\phi}) 
\defn \Exzy\{D_{\vec{\phi}}(\vec{x},\vec{y}) - D_{\vec{\phi}}(G_{\vec{\theta}}(\vec{z},\vec{y}),\vec{y})\}
\label{eq:Ladv} 
\end{align}
with respect to $\vec{\theta}$ and minimizing $-\Ladv(\vec{\theta},\vec{\phi}) + \Lgp(\vec{\phi})$ with respect to $\vec{\phi}$, where $\Lgp(\vec{\phi})$ is a gradient penalty that is used to encourage $D_{\vec{\phi}}\in L_1$ \cite{Gulrajani:NIPS:17}.
Furthemore, the expectation over $\vec{x}$ and $\vec{y}$ in \eqref{Ladv} is replaced in practice by a sample average over the training examples $\{(\vec{x}_t,\vec{y}_t)\}_{t=1}^T$.

One of the main challenges with the cGAN framework in image recovery problems is that, for each measurement example $\vec{y}_t$, there is only a single image example $\vec{x}_t$.
Thus, with the previously described training methodology, there is no incentive for the generator to produce diverse samples $G(\vec{z},\vec{y})|_{\vec{z}\sim\pz}$ for a fixed $\vec{y}$.
This can lead to the generator learning to ignore the code vector $\vec{z}$, which causes a form of ``mode collapse.''

Although issues with stability and mode collapse are also present in \emph{unconditional} GANs (uGANs) or discretely conditioned cGANs \cite{Mirza:14},
the causes are fundamentally different than in continuously conditioned cGANs like ours. 
With continuously conditioned cGANs, there is only \emph{one} example of a valid $\vec{x}_t$ for each given $\vec{y}_t$, whereas with uGANs there are many $\vec{x}_t$ and with discretely conditioned cGANs there are many $\vec{x}_t$ for each conditioning class.
As a result, most strategies that are used to combat mode-collapse in uGANs \cite{Schonfeld:CVPR:20,Karras:NIPS:20,Zhao:AAAI:21} are not well suited to cGANs.
For example, mini-batch discrimination strategies like MBSD \cite{Karras:ICLR:18}%
, where the discriminator aims to distinguish a mini-batch of true samples $\{\vec{x}_t\}$ from a mini-batch of generated samples $\{\hvec{x}_t\}$,
don't work with cGANs because the posterior statistics are very different than the prior statistics. 

To combat mode collapse in cGANs, Adler \& \"Oktem~\cite{Adler:18} proposed to use a three-input discriminator $\Dadler_{\vec{\phi}}:\mc{X}\times\mc{X}\times\mc{Y}\rightarrow \Real$ and replace $\Ladv$ from \eqref{Ladv} with the loss 
\begin{align}
\Ladler(\vec{\theta},\vec{\phi}) 
&\defn \Exzzy\big\{ 
        \tfrac{1}{2}\Dadler_{\vec{\phi}}(\vec{x},G_{\vec{\theta}}(\vec{z}_1,\vec{y}),\vec{y}) + \tfrac{1}{2}\Dadler_{\vec{\phi}}(G_{\vec{\theta}}(\vec{z}_2,\vec{y}),\vec{x},\vec{y})
        \nonumber\\&\quad
        - \Dadler_{\vec{\phi}}(G_{\vec{\theta}}(\vec{z}_1,\vec{y}),G_{\vec{\theta}}(\vec{z}_2,\vec{y}),\vec{y})
         \big\} 
\label{eq:Ladler},
\end{align}
which rewards variation between the first and second inputs to $\Dadler_{\vec{\phi}}$.
They then proved that minimizing $\Ladler$ in place of $\Ladv$ does not compromise the Wasserstein cGAN objective, i.e., $\arg\min_{\vec{\theta}} \Ladler(\vec{\theta},\vec{\phi}) = \arg\min_{\vec{\theta}} \Ladv(\vec{\theta},\vec{\phi})$.
As we show in \secref{experiments}, this approach does prevent complete mode collapse, but it leaves much room for improvement.

\section{Proposed method} \label{sec:proposed}

\subsection{Proposed regularization: supervised-\texorpdfstring{$\ell_1$}{L1} plus SD reward} \label{sec:L1}

We now propose a novel cGAN regularization framework.
To train the generator, we propose to solve
\begin{align}\textstyle
\arg\min_{\vec{\theta}} \{\betaadv\Ladv(\vec{\theta},\vec{\phi})+\LonestdPt(\vec{\theta},\betastd)\}
\end{align}
with appropriately chosen $\betaadv,\betastd>0$ and $P\train\geq 2$, where the regularizer
\begin{align}\textstyle
\LonestdPt(\vec{\theta},\betastd) \defn \LonePt(\vec{\theta}) - \betastd\LstdPt(\vec{\theta})
\label{eq:LonestdP}
\end{align}
is constructed from the $P\train$-sample supervised-$\ell_1$ loss and standard-deviation (SD) reward terms 
\begin{align}
\LonePt(\vec{\theta})
&\defn \textstyle \ExzPy\big\{ \|\vec{x}-\hvec{x}\avgPt\|_1\big\}
\label{eq:LoneP} \\
\LstdPt(\vec{\theta})
&\defn \textstyle \sqrt{\frac{\pi}{2P\train(P\train-1)}} \sum_{i=1}^{P\train} \EzPy\big\{ \|\hvec{x}_i-\hvec{x}\avgPt\|_1 \big\}
\label{eq:LstdP} ,
\end{align}
and where $\{\hvec{x}_i\}$ denote the generated samples and $\hvec{x}\avgP$ their $P$-sample average:
\begin{align}
\hvec{x}_i \defn G_{\vec{\theta}}(\vec{z}_i,\vec{y}),
\qquad
\hvec{x}\avgP \defn \textstyle \frac{1}{P} \sum_{i=1}^P \hvec{x}_i
\label{eq:xhat}. 
\end{align}
The use of supervised-$\ell_1$ loss and SD reward in \eqref{LonestdP} is not heuristic.
As shown in \propref{L1std}, it encourages the samples $\{\hvec{x}_i\}$ to match the true posterior in both mean and covariance.

\begin{proposition} \label{prop:L1std}
Suppose $P\train\geq 2$ and $\vec{\theta}$ has complete control over the $\vec{y}$-conditional mean and covariance of $\hvec{x}_i$.
Then the parameters
$\vec{\theta}_* = \arg\min_{\vec{\theta}} \LonestdPt(\vec{\theta},\betastdG)$ with
\begin{align}
\textstyle \betastdG \defn \sqrt{\frac{2}{\pi P\train(P\train+1)}}
\label{eq:betastdG}
\end{align}
yield generated statistics
\begin{subequations}\label{eq:goal}
\begin{align}
\Ezigy\{\hvec{x}_i(\vec{\theta}_*)|\vec{y}\} &= \Exgy\{\vec{x}|\vec{y}\} = \hvec{x}\mmse \\
\Covzigy\{\hvec{x}_i(\vec{\theta}_*)|\vec{y}\} &= \Covxgy\{\vec{x}|\vec{y}\}
\end{align}
\end{subequations}
when the elements of $\hvec{x}_i$ and $\vec{x}$ are independent Gaussian conditioned on $\vec{y}$.
Thus, minimizing $\LonestdPt$ encourages the $\vec{y}$-conditional mean and covariance of $\hvec{x}_i$ to match those of the true $\vec{x}$.  
\end{proposition}

See \appref{L1std} for a proof.
In imaging applications, $\hvec{x}_i$ and $\vec{x}$ may not be independent Gaussian conditioned on $\vec{y}$, and so the value of $\betastd$ in \eqref{betastdG} may not be appropriate. 
Thus we propose a method to automatically tune $\betastd$ in \secref{betastd}. 

\begin{figure*}[t]
\centering
\includegraphics[width=0.245\linewidth]{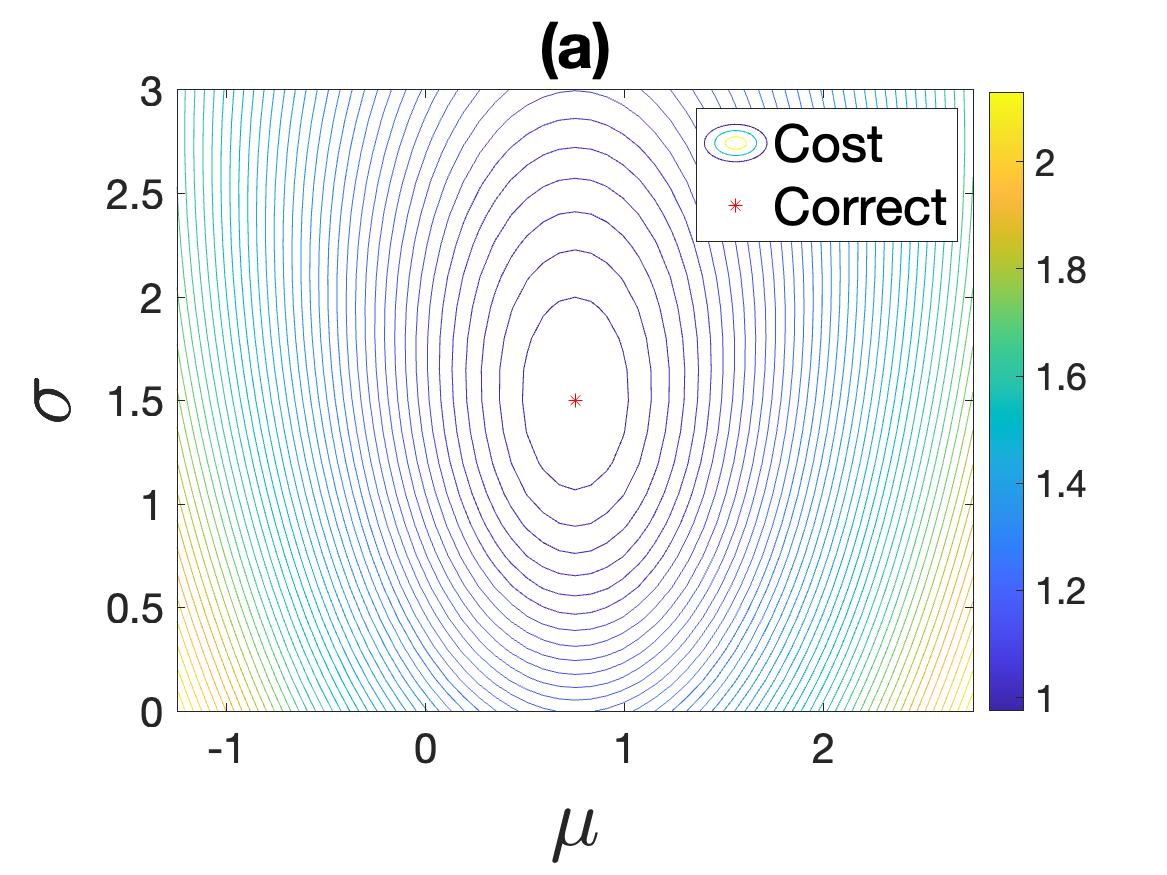}
\includegraphics[width=0.245\linewidth]{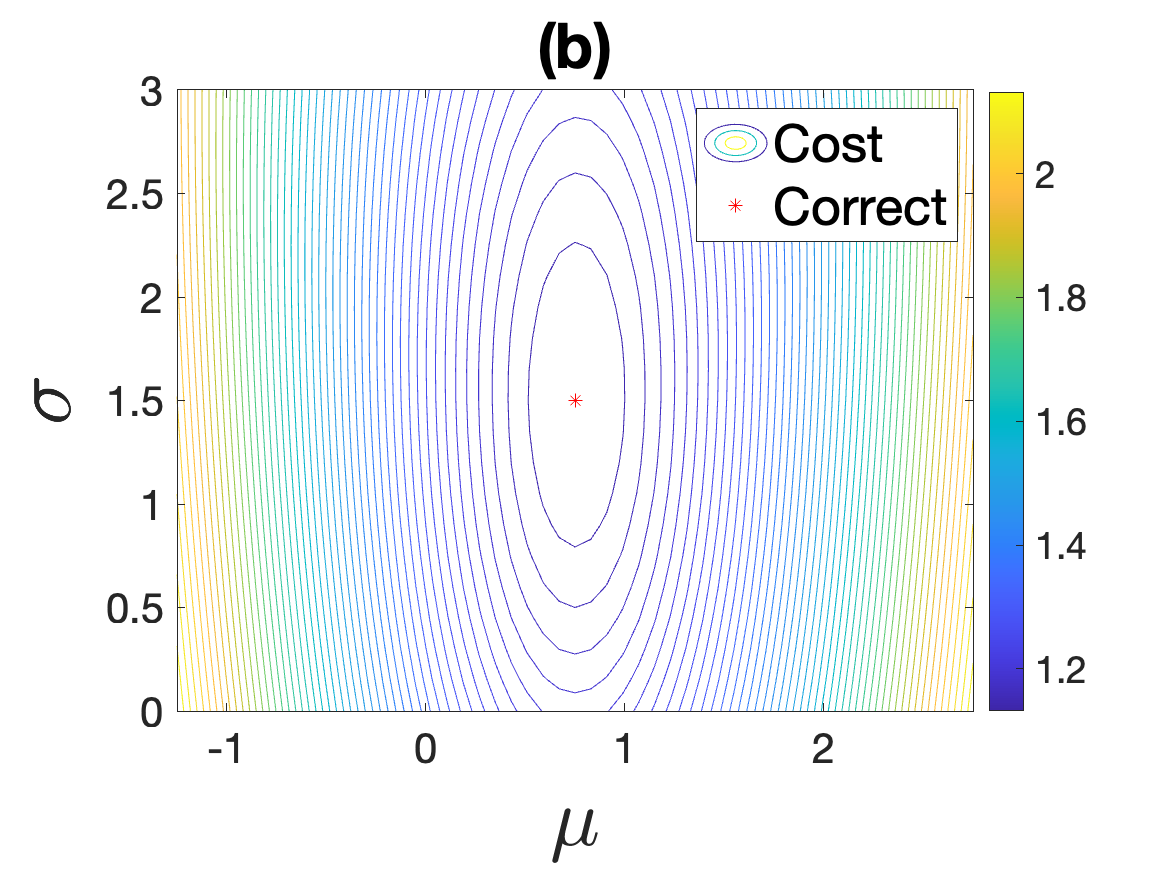}
\includegraphics[width=0.245\linewidth]{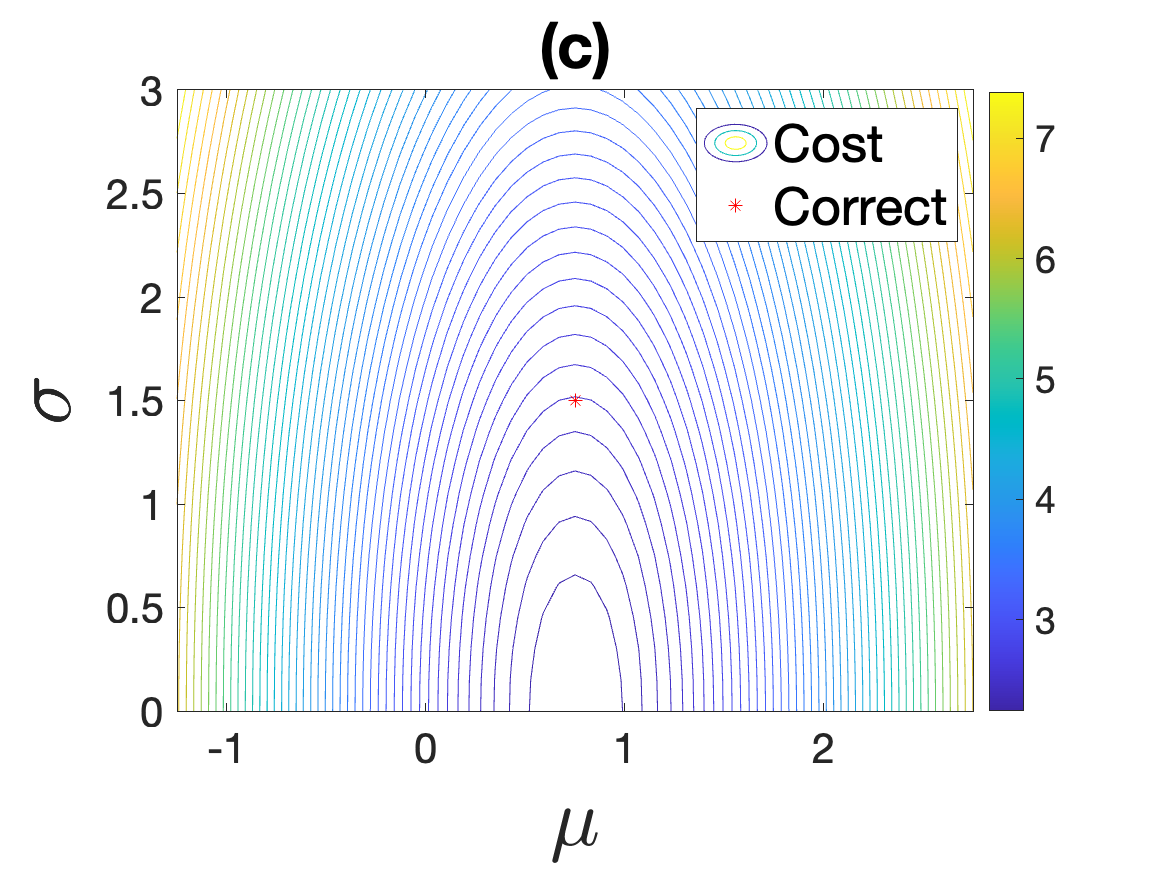}
\includegraphics[width=0.245\linewidth]{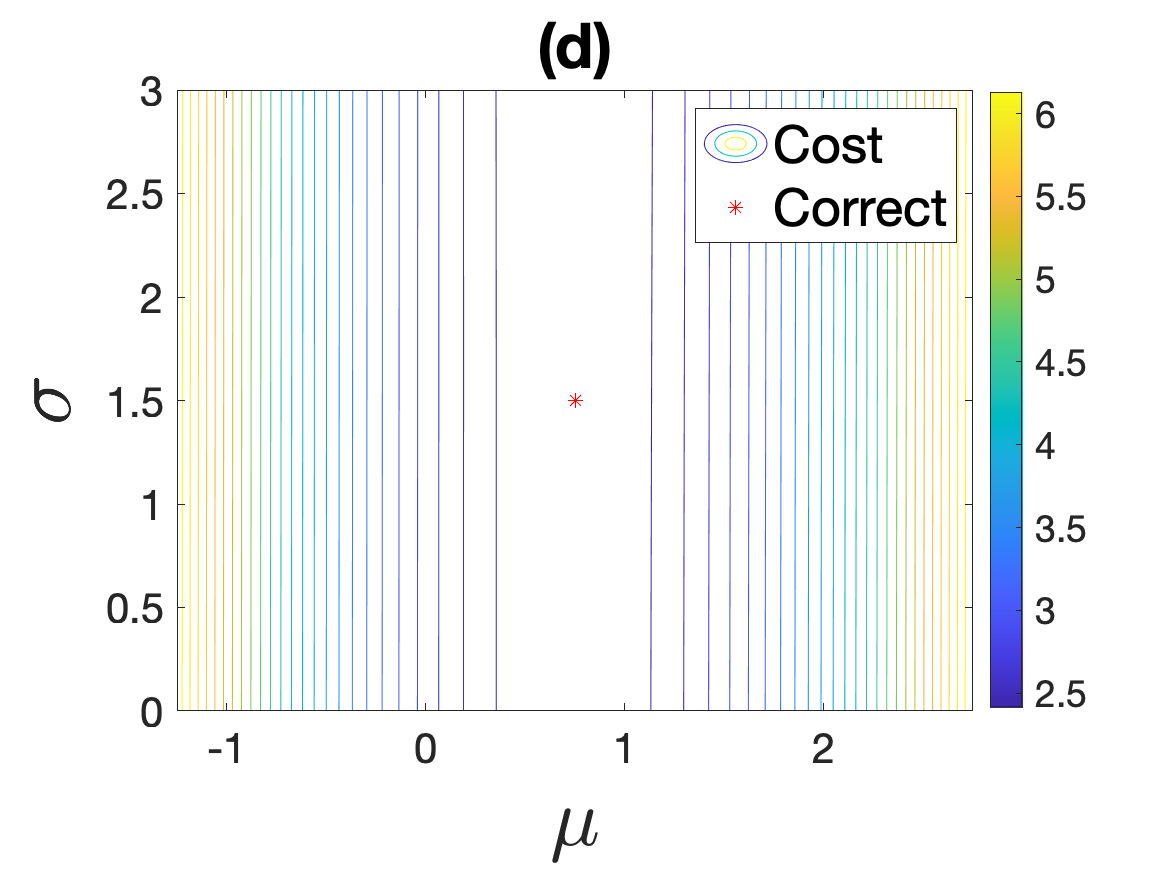}\\[-3mm]
\caption{The contours show the regularizer value versus $\vec{\theta}\!=\![\mu,\sigma]\tran$ for four different regularizers:
(a) supervised-$\ell_1$ plus SD reward with $\betastd\!=\!\betastdG$ at $P\train\!=\!2$,
(b) supervised-$\ell_1$ plus SD reward with $\betastd\!=\!\betastdG$ at $P\train\!=\!8$,
(c) supervised-$\ell_2$ at $P\train\!=\!8$, 
and
(d) supervised-$\ell_2$ plus variance reward at $P\train\!=\!8$.
The red star shows the true posterior parameters $[\mu_0,\sigma_0]\tran$.
}
\label{fig:illustration}
\end{figure*}

\Figref{illustration} shows a toy example with 
parameters $\vec{\theta}=[\mu,\sigma]\tran$,
generator $G_{\vec{\theta}}(z,y)=\mu+\sigma z$, 
and $z\sim\mc{N}(0,1)$, giving generated posterior
$\pxhgy(x|y)=\mc{N}(x;\mu,\sigma^2)$.
Assuming the true $\pxgy(x|y)=\mc{N}(x;\mu_0,\sigma_0^2)$,
Figs.~\ref{fig:illustration}(a)-(b) show that, by minimizing the proposed $\LonestdPt(\vec{\theta},\betastdG)$ regularization over $\vec{\theta}=[\mu,\sigma]\tran$ for any $P\train\geq 2$, we recover the true $\vec{\theta}_0=[\mu_0,\sigma_0]\tran$.
They also show that the cost function steepens as $P\train$ decreases, with agrees with our empirical finding that $P\train=2$ tends to work best in practice.

We note that regularizing a cGAN with supervised-$\ell_1$ loss alone is not new; see, e.g., \cite{Isola:CVPR:17}.
In fact, the use of supervised-$\ell_1$ loss is often preferred over $\ell_2$ in image recovery because it results in sharper, more visually pleasing results \cite{Zhao:TCI:17}. 
But regularizing a cGAN using supervised-$\ell_1$ loss \emph{alone} can push the generator towards mode collapse, for reasons described below.
For example, in \cite{Isola:CVPR:17}, $\ell_1$-induced mode collapse led the authors to use dropout to induce generator variation, instead of random $\vec{z}_i$.


\paragraph{Why not supervised-\texorpdfstring{$\ell_2$}{L2} regularization?} 
One may wonder: Why regularize using supervised-$\ell_1$ loss plus an SD reward in \eqref{LonestdP} and not a more conventional choice like supervised-$\ell_2$ loss plus a variance reward, or even supervised-$\ell_2$ loss alone?
We start by discussing the latter.

The use of supervised-$\ell_2$ regularization in a cGAN can be found in \cite{Isola:CVPR:17}.
In this case, to train the generator, one would aim to solve $\arg\min_{\vec{\theta}} \{\Ladv(\vec{\theta},\vec{\phi}) + \lambda \Ltwo(\vec{\theta})\}$ with some $\lambda>0$ and
\begin{align}
\Ltwo(\vec{\theta})
&\defn \Exy\big\{ \|\vec{x}-\Ez\{G_{\vec{\theta}}(\vec{z},\vec{y})\}\|^2_2 \big\}
\label{eq:Ltwo} .
\end{align}
Ohayon et al.~\cite{Ohayon:ICCVW:21} revived this idea for the explicit purpose of fighting mode collapse.
In practice, however, the $\Ez$ term in \eqref{Ltwo} must be implemented by a finite-sample average, giving
\begin{align}
\LtwoPt(\vec{\theta})
&\defn \textstyle \ExzPy\big\{ \big\|\vec{x}-\frac{1}{P\train}\sum_{i=1}^{P\train} G_{\vec{\theta}}(\vec{z}_i,\vec{y}) \big\|^2_2 \big\} 
\label{eq:LtwoP} 
\end{align}
for some $P\train\geq 2$.
For example, Ohayon's implementation \cite{Ohayon:github:21} used $P\train=8$.
As we show in \propref{L2P}, $\LtwoPt$ \emph{induces} mode collapse rather than prevents it. 

\begin{proposition} \label{prop:L2P}
Say $P\train$ is finite and $\vec{\theta}$ has complete control over the $\vec{y}$-conditional mean and covariance of $\hvec{x}_i$.
Then the parameters
$\vec{\theta}_* = \arg\min_{\vec{\theta}} \LtwoPt(\vec{\theta})$ 
yield generated statistics
\begin{subequations}
\begin{align}
\Ezigy\{\hvec{x}_i(\vec{\theta}_*)|\vec{y}\} &= \Exgy\{\vec{x}|\vec{y}\} = \hvec{x}\mmse \\
\Covzigy\{\hvec{x}_i(\vec{\theta}_*)|\vec{y}\} &= \vec{0} .
\end{align}
\end{subequations}
Thus, minimizing $\LtwoPt$ encourages mode collapse.
\end{proposition}

The proof (see \appref{L2P}) follows from the bias-variance decomposition of \eqref{LtwoP}, i.e.,
\begin{align}
&\lefteqn{\LtwoPt(\vec{\theta})}\nonumber\\
&= \textstyle \Ey\big\{ 
    \|\hvec{x}\mmse-\Ezigy\{\hvec{x}_i(\vec{\theta})|\vec{y}\}\|_2^2 
    + \tfrac{1}{P\train} 
      \tr [ \Covzigy\{\hvec{x}_i(\vec{\theta}) | \vec{y}\} ]
    + \Exgy\{\|\vec{e}\mmse\|_2^2 | \vec{y}\} 
    \big\}
\label{eq:LtwoPerrt} ,
\end{align}
where $\vec{e}\mmse\defn \vec{x}-\vec{x}\mmse$ is the MMSE error.
\Figref{illustration}(c) shows that $\LtwoPt$ regularization causes mode collapse in the toy example, and \secref{MRIexperiments} shows that it causes mode collapse in MRI. 


\paragraph{Why not supervised $\ell_2$ plus a variance reward?} 
To mitigate $\LtwoPt$'s incentive for mode-collapse, the second term in \eqref{LtwoPerrt} could be canceled using a variance reward, giving 
\begin{align}
\LtwovarPt(\vec{\theta})
&\defn \textstyle \LtwoPt(\vec{\theta}) - \frac{1}{P\train}\LvarPt(\vec{\theta})
\label{eq:LtwovarP} \\
\text{with~}\LvarPt(\vec{\theta})
&\defn \textstyle \frac{1}{P\train-1}\sum_{i=1}^{P\train} \EzPy\{\|\hvec{x}_i(\vec{\theta})-\hvec{x}\avgP(\vec{\theta})\|_2^2\}
\label{eq:LvarP} .
\end{align}
since \appref{LvarPerr} shows that $\LvarPt(\vec{\theta})$ is an unbiased estimator of the posterior trace-covariance:
\begin{align}
\LvarPt(\vec{\theta})
&= \textstyle \Ey \{ \tr [ \Covzigy\{\hvec{x}_i(\vec{\theta}) | \vec{y}\} ] \}
\text{~~for any~} P\train\geq 2
\label{eq:LvarPerr}.
\end{align}
However, the resulting $\LtwovarPt$ regularizer in \eqref{LtwovarP} does not encourage the generated covariance to match the \emph{true} posterior covariance, unlike the proposed $\LonestdPt$ regularizer in \eqref{LonestdP} (recall \propref{L1std}).
For the toy example, this behavior is visible in \figref{illustration}(d).

\subsection{Auto-tuning of SD reward weight \texorpdfstring{$\betastd$}{betastd}} \label{sec:betastd}

\begin{figure*}
    \centering
    \footnotesize
    \def\tabwid{19.5mm}
    \begin{tabular}{>{\centering}p{\tabwid}
                    >{\centering}p{\tabwid}
                    >{\centering}p{\tabwid}
                    >{\centering}p{\tabwid}
                    >{\centering}p{\tabwid}
                    >{\centering}p{\tabwid}}
        $\betastd=0$ & $\betastd=\betastdG$ & $\betastd=1.2\betastdG$ & $\betastd=1.4\betastdG$ & $\betastd=1.6\betastdG$ & $\betastd=1.8\betastdG$
    \end{tabular}\\
    \includegraphics[width=\linewidth,trim=110 10 90 25,clip]{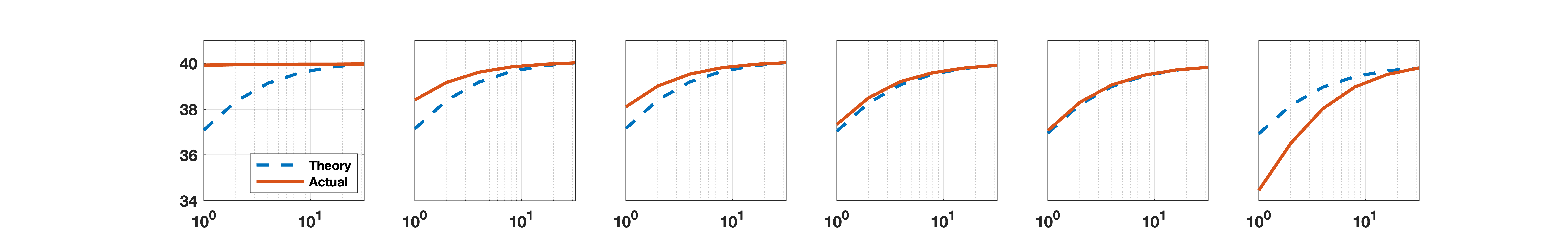}\\
    {\sf Number of averaged outputs, $P$, on a log scale}
    \caption{Example PSNR of $\hvec{x}\avgP$ versus $P$, the number of averaged outputs, for several training $\betastd$ and MRI recovery at $R=4$. Also shown is the theoretical behavior for true-posterior samples.}
    \label{fig:betastd_ablation}
\end{figure*}

We now propose a method to auto-tune $\betastd$ in \eqref{LonestdP} for a given training dataset.
Our approach is based on the observation that larger values of $\betastd$ tend to yield samples $\hvec{x}_i$ with more variation.
But more variation is not necessarily better; we want samples with the correct amount of variation.
To assess the correct amount of variation, we compare the expected $\ell_2$ error of the $P$-sample average $\hvec{x}\avgP$ to that of $\hvec{x}\avgone$.
When $\{\hvec{x}_i\}$ are true-posterior samples, these errors follow a particular relationship, as established by \propref{ratio} below (see \appref{ratio} for a proof).
\begin{proposition} \label{prop:ratio}
Say $\hvec{x}_i\sim \pxgy(\cdot|\vec{y})$ are independent samples of the true posterior and, for any $P\geq 1$, their $P$-sample average is $\hvec{x}\avgP\defn \frac{1}{P}\sum_{i=1}^P\hvec{x}_i$.
Then
\begin{align}
\mc{E}_P 
&\defn \E\{\|\hvec{x}\avgP-\vec{x}\|_2^2 | \vec{y}\}
= \tfrac{P+1}{P} \mc{E}\mmse 
\text{,\quad which implies\quad}
\tfrac{\mc{E}_1}{\mc{E}_P} = \tfrac{2P}{P+1} 
\label{eq:Eratio}.
\end{align}
\end{proposition}

Experimentally we find that $\mc{E}_1/\mc{E}_P$ grows with the SD reward weight $\betastd$.
(See \figref{betastd_ablation}.)
Thus, we propose to adjust $\betastd$ so that the observed SNR-gain ratio $\mc{E}_1/\mc{E}_{P\val}$ matches the correct ratio $(2P\val)/(P\val+1)$ from \eqref{Eratio}, for some $P\val\geq 2$ that does not need to match $P\train$.
(We use $P\val=8$ in \secref{experiments}.)
In particular, at each training epoch $\tau$, we approximate $\mc{E}_{P\val}$ and $\mc{E}_1$ 
as follows:
\begin{align}
\hat{\mc{E}}_{P\val,\tau} 
&\defn \textstyle \frac{1}{V}\sum_{v=1}^V \|\frac{1}{P\val}\sum_{i=1}^{P\val} G_{\vec{\theta}_\tau}(\vec{z}_{i,v},\vec{y}_v)-\vec{x}_v\|_2^2 \\
\hat{\mc{E}}_{1,\tau} 
&\defn \textstyle \frac{1}{V}\sum_{v=1}^V \|G_{\vec{\theta}_\tau}(\vec{z}_{1,v},\vec{y}_v)-\vec{x}_v\|_2^2 ,
\end{align}
with validation set $\{(\vec{x}_v,\vec{y}_v)\}_{v=1}^V$ and 
i.i.d.\ codes $\{\vec{z}_{i,v}\}_{i=1}^{P\val}$.
We update $\betastd$ using gradient descent:
\begin{align}
\beta_{\mathsf{SD},\tau+1}
&= \textstyle \beta_{\mathsf{SD},\tau} - \mustd \cdot \big( 
[\hat{\mc{E}}_{1,\tau}/\hat{\mc{E}}_{P\val,\tau}]_{\text{dB}} 
- [2P\val/(P\val+1)]_{\text{dB}} 
\big) \betastdG
\text{~~for~~}\tau=0,1,2,\dots
\label{eq:betastd}
\end{align}
with 
$\beta_{\mathsf{SD},0} = \betastdG$,
some $\mustd>0$, 
and $[x]_{\text{dB}}\defn 10\log_{10}(x)$.

\section{Numerical experiments} \label{sec:experiments}

\subsection{Conditional Fr\'echet inception distance} \label{sec:cfid}

As previously stated, our goal is to train a generator $G_{\vec{\theta}}$ so that, for typical fixed values of $\vec{y}$, the generated distribution $\pxhgy(\cdot|\vec{y})$ matches the true posterior $\pxgy(\cdot|\vec{y})$.
It is essential to have a quantitative metric for evaluating performance with respect to this goal.
For example, it is not enough that the generated samples are ``accurate'' in the sense of being close to the ground truth, nor is it enough that they are ``diverse'' according to some heuristically chosen metric.

We quantify posterior-approximation quality using the conditional Fr\'{e}chet inception distance (CFID) \cite{Soloveitchik:21}, a computationally efficient approximation to the conditional Wasserstein distance
\begin{align}
\CWD \defn \Ey\{W_2(\pxgy(\cdot,\vec{y}),\pxhgy(\cdot,\vec{y}))\} 
\label{eq:CWD} .
\end{align}
In \eqref{CWD}, $W_2(\pa,\pb)$ denotes the Wasserstein-2 distance between distributions $\pa$ and $\pb$, defined as
\begin{align}
&W_2(\pa,\pb) 
\defn \min_{\pab\in \Pi(\pa,\pb)} \Eab\{\|\vec{a}-\vec{b}\|^2_2\} 
\label{eq:Pi} ,
\end{align}
where $\Pi(\pa,\pb) \defn \textstyle \big\{\pab : \pa=\int \pab\deriv \vec{b} \text{~and~} \pb=\int \pab \deriv \vec{a} \big\} $ denotes the set of joint distributions $\pab$ with prescribed marginals $\pa$ and $\pb$.
Similar to how FID \cite{Heusel:NIPS:17}---a popular uGAN metric---is computed, CFID approximates CWD \eqref{CWD} as follows:
i) the random vectors $\vec{x}$, $\hvec{x}$, and $\vec{y}$ are replaced by low-dimensional embeddings $\uvec{x}$, $\huvec{x}$, and $\uvec{y}$, generated by the convolutional layers of a deep network, and
ii) the embedding distributions $\pxugyu$ and $\pxuhgyu$ are approximated by multivariate Gaussians.
More details on CFID are given in \appref{cfid}.

\subsection{MRI experiments} \label{sec:MRIexperiments}
We consider parallel MRI recovery, where the goal is to recover a complex-valued multicoil image $\vec{x}$ from zero-filled measurements $\vec{y}$ (see \appref{MRI} for details).

\textbf{Data.~}
For training data $\{\vec{x}_t\}$, we used the first 8 slices of all fastMRI \cite{Zbontar:18} T2 brain training volumes with at least 8 coils, cropping to $384 \times 384$ pixels and compressing to 8 virtual coils \cite{Zhang:MRM:13}, yielding 12\,200 training images.
Using the same procedure, 2\,376 testing and 784 validation images were obtained from the fastMRI T2 brain testing volumes. 
From the 2\,376 testing images, 72 were randomly selected to evaluate the Langevin technique \cite{Jalal:NIPS:21} in order to limit its sample generation to 6 days.
To create the measurement $\vec{y}_t$, we transformed $\vec{x}_t$ to the Fourier domain, sampled using pseudo-random GRO patterns \cite{Joshi:22} at acceleration $R=4$ and $R=8$, and Fourier-transformed the zero-filled k-space measurements back to the (complex, multicoil) image domain.

\textbf{Architecture.~} 
We use a UNet \cite{Ronneberger:MICCAI:15} for our generator and a standard CNN for our generator, along with data-consistency as in \appref{consistency}.
Architecture and training details are given in \appref{implementation}. 

\textbf{Competitors.~} 
We compare our cGAN to the Adler and \"Oktem's cGAN \cite{Adler:18}, Ohayon et al.'s cGAN \cite{Ohayon:ICCVW:21}, Jalal et al.'s fastMRI Langevin approach \cite{Jalal:NIPS:21}, and Sriram et al.'s E2E-VarNet \cite{Sriram:MICCAI:20}.
The cGAN from \cite{Adler:18} uses generator loss $\betaadv\Ladler(\vec{\theta},\vec{\phi})$ and discriminator loss $-\Ladler(\vec{\theta},\vec{\phi})+ \alpha_1 \Lgp(\vec{\phi}) + \alpha_2 \Ldrift(\vec{\phi})$, 
while the cGAN from \cite{Ohayon:ICCVW:21} uses generator loss $\betaadv\Ladv(\vec{\theta},\vec{\phi}) + \LtwoP(\vec{\theta})$ and discriminator loss $-\Ladv(\vec{\theta},\vec{\phi})+ \alpha_1 \Lgp(\vec{\phi}) + \alpha_2 \Ldrift(\vec{\phi})$.
Each used the value of $\betaadv$ specified in the original paper. 
All cGANs used the same generator and discriminator architectures, except that \cite{Adler:18} used extra discriminator input channels to facilitate the 3-input loss \eqref{Ladler}.
For the fastMRI Langevin approach \cite{Jalal:NIPS:21}, we did not modify the authors' implementation in \cite{Jalal:github:21} except to use the GRO sampling mask.
For the E2E-VarNet \cite{Sriram:MICCAI:20}, we use the same training procedure and hyperparameters outlined in \cite{Jalal:NIPS:21} except that we use the GRO sampling mask.

\textbf{Testing.~} 
To evaluate performance, we converted the multicoil outputs $\hvec{x}_i$ to complex-valued images using SENSE-based coil combining \cite{Prussmann:MRM:99} with ESPIRiT-estimated \cite{Uecker:MRM:14} coil sensitivity maps, as described in \appref{MRI}.
Magnitude images were used to compute performance metrics.


\begin{table*}[h]
  \caption{Average MRI results at $R\in\{4,8\}$. 
    CFID$^1$, FID, and APSD used $72$ test samples and $P\!=\!32$, 
    CFID$^2$ used $2\,376$ test samples and $P\!=\!8$, and 
    CFID$^3$ used all $14\,576$ samples and $P\!=\!1$}
  \vspace{0.1in}
  \label{tab:mriresults}
  \centering
  \resizebox{\textwidth}{!}{%
  \begin{tabular}{llllllllllllll}
    \toprule
    & \multicolumn{6}{c}{$R=4$} & \multicolumn{6}{c}{$R=8$}\\
    \cmidrule(lr){2-7} \cmidrule(lr){8-13}
    Model & 
    CFID$^1\!\!\downarrow$ & CFID$^2\!\!\downarrow$ & CFID$^3\!\!\downarrow$ & FID$\downarrow$ & APSD & Time (4)$\downarrow$ &
    CFID$^1\!\!\downarrow$ & CFID$^2\!\!\downarrow$ & CFID$^3\!\!\downarrow$ & FID$\downarrow$ & APSD & Time (4)$\downarrow$ \\
    \midrule
    E2E-VarNet (Sriram et al.\ \cite{Sriram:MICCAI:20}) 
        & 7.47 & 6.99 & 6.61 & 8.84 & 0.0 & 310ms 
        & 7.82 & 6.81 & 6.31 & 8.40 & 0.0 & 316ms\\
    Langevin (Jalal et al.\ \cite{Jalal:NIPS:21}) 
        & 5.29 & - & - & 6.12 & 5.9e-6 & 14 min
        & 7.34 & - & - & 14.32  & 7.6e-6 & 14 min\\
    cGAN (Adler \& \"Oktem \cite{Adler:18}) 
        & 6.39 & 4.27 & 3.82 & 5.25 & 3.9e-6 & \textbf{217 ms}
        & 10.10 & 6.30 & 5.72 & 10.77  & 7.7e-6 & \textbf{217 ms}\\
    cGAN (Ohayon et al.\ \cite{Ohayon:ICCVW:21}) 
        & 4.06 & 3.27 & 2.95 & 6.45 & 7.2e-8 & \textbf{217 ms}
        & 6.04 & 4.59 & 4.27 & 11.05  & 7.7e-7 & \textbf{217 ms}\\
    cGAN (Ours) & \textbf{3.10} 
        & \textbf{1.54} & \textbf{1.29} & \textbf{3.75} & 3.8e-6 & \textbf{217 ms}
        & \textbf{4.87} & \textbf{2.23} & \textbf{1.79} & \textbf{7.72}  & 7.6e-6 & \textbf{217 ms}\\
    \bottomrule
  \end{tabular}%
  }
\end{table*}

\textbf{Results.~} 
\tabref{mriresults} shows CFID, FID, APSD $\defn(\frac{1}{NP}\sum_{i=1}^{P} \|\hvec{x}\avgP-\hvec{x}_i\|^2)^{1/2}$, and 4-sample generation time at $R\in\{4,8\}$.
(C)FID was computed using VGG-16 (not Inception-v3) to better align with radiologists' perceptions \cite{Kastryulin:22}.
As previously described, the Langevin method was evaluated using only 72 test images.
Because CFID is biased at small sample sizes \cite{Soloveitchik:21}, we re-evaluated the other methods using all $2\,376$ test images, and again using all $14\,576$ training and test images. 
\tabref{mriresults} shows that our approach gave significantly better CFID and FID than the competitors.
Also, the APSD of Ohayon et al.'s cGAN was an order-of-magnitude smaller than the others, indicating mode collapse.
The cGANs generated samples $3\,800$ times faster than the Langevin approach from \cite{Jalal:NIPS:21}.

Tables \ref{tab:versusPr4} and \ref{tab:versusPr8} show PSNR, SSIM, LPIPS \cite{Zhang:CVPR:18}, and DISTS \cite{Ding:TPAMI:20} for the $P$-sample average $\hvec{x}\avgP$ at $P\in\{1,2,4,8,16,32\}$ and $R\in\{4,8\}$, respectively.
While the E2E-VarNet achieves the best PSNR at $R\in\{4,8\}$ and the best SSIM at $R=4$, the proposed cGAN achieves the best LPIPS and DISTS performances at $R\in\{4,8\}$ when $P=2$ and the best SSIM at $R=8$ when $P=8$. 
The $P$ dependence can be explained by the perception-distortion tradeoff \cite{Blau:CVPR:18}: as $P$ increases, $\hvec{x}\avgP$ transitions from better perceptual quality to lower $\ell_2$ distortion. 
PSNR favors $P\rightarrow\infty$ (e.g., $\ell_2$ optimality) while the other metrics favor particular combinations of perceptual quality and distortion.
The plots in Appendices~\ref{app:mriplotsR4} and \ref{app:mriplotsR8} show zoomed-in versions of $\hvec{x}\avgP$ that visually demonstrate the perception-distortion tradeoff at $P\in\{1,2,4,32\}$: smaller $P$ yield sharper images with more variability from the ground truth, while larger $P$ yield smoother reconstructions. 

\begin{table*}[t]
  \caption{Average PSNR, SSIM, LPIPS, and DISTS of $\hvec{x}\avgP$ versus $P$ for $R=4$ MRI}
  \vspace{0.1in}
  \label{tab:versusPr4}
  \centering
  \resizebox{1.0\textwidth}{!}{%
  \begin{tabular}{lcccccccccccc}
    \toprule
    & \multicolumn{6}{c}{PSNR$\uparrow$} 
        & \multicolumn{6}{c}{SSIM$\uparrow$}\\ 
    \cmidrule(lr){2-7} \cmidrule(lr){8-13}
    Model & $P\!=$1 & $P\!=$2 & $P\!=$4 & $P\!=$8 & $P\!=$16 & $P\!=$32
          & $P\!=$1 & $P\!=$2 & $P\!=$4 & $P\!=$8 & $P\!=$16 & $P\!=$32\\
    \midrule
    E2E-VarNet (Sriram et al. \cite{Sriram:MICCAI:20}) & \bf 39.93 & - & - & - & - & - & \bf 0.9641 & - & - & - & - & -\\
    Langevin (Jalal et al. \cite{Jalal:NIPS:21}) 
          & 36.04 & 37.02 & 37.65 & 37.99 & 38.17 & 38.27 &  0.8989 & 0.9138 & 0.9218 & 0.9260 & 0.9281 & 0.9292\\
    cGAN (Adler \& \"Oktem \cite{Adler:18}) 
          & 35.63 & 36.64 & 37.24 & 37.56 & 37.73 & 37.82 &  0.9330 & 0.9445 & 0.9478 & 0.9480 & 0.9477 & 0.9473\\
    cGAN (Ohayon et al. \cite{Ohayon:ICCVW:21}) 
          & 39.44 & 39.46 & 39.46 & 39.47 & 39.47 & 39.47 &  0.9558 & 0.9546 & 0.9539 & 0.9535 & 0.9533 & 0.9532\\
    cGAN (Ours) 
          & 36.96 & 38.14 & 38.84 & 39.24 & 39.44 & 39.55 &  0.9440 & 0.9526 & 0.9544 & 0.9542 & 0.9537 & 0.9533\\
    & \multicolumn{6}{c}{LPIPS$\downarrow$} 
        & \multicolumn{6}{c}{DISTS$\downarrow$}\\ 
    \cmidrule(lr){2-7} \cmidrule(lr){8-13}
    Model & $P\!=$1 & $P\!=$2 & $P\!=$4 & $P\!=$8 & $P\!=$16 & $P\!=$32
          & $P\!=$1 & $P\!=$2 & $P\!=$4 & $P\!=$8 & $P\!=$16 & $P\!=$32\\
    \midrule
    E2E-VarNet (Sriram et al. \cite{Sriram:MICCAI:20}) & 0.0316 & - & - & - & - & - & 0.0859 & - & - & - & - & -\\
    Langevin (Jalal et al. \cite{Jalal:NIPS:21}) 
          & 0.0545 & 0.0394 & 0.0336 & 0.0320 & 0.0317 & 0.0316 &  0.1116 & 0.0921 & 0.0828 & 0.0793 & 0.0781 & 0.0777\\
    cGAN (Adler \& \"Oktem \cite{Adler:18}) 
          & 0.0285 & 0.0255 & 0.0273 & 0.0298 & 0.0316 & 0.0327 &  0.0972 & 0.0857 & 0.0878 & 0.0930 & 0.0967 & 0.0990\\
    cGAN (Ohayon et al. \cite{Ohayon:ICCVW:21}) 
          & 0.0245 & 0.0247 & 0.0248 & 0.0249 & 0.0249 & 0.0249 &  0.0767 & 0.0790 & 0.0801 & 0.0807 & 0.0810 & 0.0811\\
    cGAN (Ours) 
          & 0.0175 & \bf 0.0164 & 0.0188 & 0.0216 & 0.0235 & 0.0245 &  \bf 0.0546 & 0.0563 & 0.0667 & 0.0755 & 0.0809 & 0.0837\\
    \bottomrule
  \end{tabular}%
  }
\end{table*}

\begin{table*}[t]
  \caption{Average PSNR, SSIM, LPIPS, and DISTS of $\hvec{x}\avgP$ versus $P$ for $R=8$ MRI}
  \vspace{0.1in}
  \label{tab:versusPr8}
  \centering
  \resizebox{1.0\textwidth}{!}{%
  \begin{tabular}{lcccccccccccc}
    \toprule
    & \multicolumn{6}{c}{PSNR$\uparrow$} 
        & \multicolumn{6}{c}{SSIM$\uparrow$}\\ 
    \cmidrule(lr){2-7} \cmidrule(lr){8-13}
    Model & $P\!=$1 & $P\!=$2 & $P\!=$4 & $P\!=$8 & $P\!=$16 & $P\!=$32
          & $P\!=$1 & $P\!=$2 & $P\!=$4 & $P\!=$8 & $P\!=$16 & $P\!=$32\\
    \midrule
    E2E-VarNet (Sriram et al.\ \cite{Sriram:MICCAI:20}) & \bf 36.49 & - & - & - & - & - & 0.9220 & - & - & - & - & -\\
    Langevin (Jalal et al. \cite{Jalal:NIPS:21}) 
          & 32.17 & 32.83 & 33.45 & 33.74 & 33.83 & 33.90
          & 0.8725 & 0.8919 & 0.9031 & 0.9091 & 0.9120 & 0.9137\\
    cGAN (Adler \& \"Oktem \cite{Adler:18}) 
          & 31.31 & 32.31 & 32.92 & 33.26 & 33.42 & 33.51
          & 0.8865 & 0.9045 & 0.9103 & 0.9111 & 0.9102 & 0.9095\\
    cGAN (Ohayon et al. \cite{Ohayon:ICCVW:21}) 
          & 34.89 & 34.90 & 34.90 & 34.90 & 34.91 & 34.92
          & 0.9222 & 0.9217 & 0.9213 & 0.9211 & 0.9211 & 0.9210\\
    cGAN (Ours) 
          & 32.32 & 33.67 & 34.53 & 35.01 & 35.27 & 35.42
          & 0.9030 & 0.9199 & 0.9252 & \bf 0.9257 & 0.9251 & 0.9246\\
    & \multicolumn{6}{c}{LPIPS$\downarrow$} 
        & \multicolumn{6}{c}{DISTS$\downarrow$}\\ 
    \cmidrule(lr){2-7} \cmidrule(lr){8-13}
    Model & $P\!=$1 & $P\!=$2 & $P\!=$4 & $P\!=$8 & $P\!=$16 & $P\!=$32
          & $P\!=$1 & $P\!=$2 & $P\!=$4 & $P\!=$8 & $P\!=$16 & $P\!=$32\\
    \midrule
    E2E-VarNet (Sriram et al.\ \cite{Sriram:MICCAI:20}) & 0.0575 & - & - & - & - & - & 0.1253 & - & - & - & - & -\\
    Langevin (Jalal et al. \cite{Jalal:NIPS:21}) 
          & 0.0769 & 0.0619 & 0.0579 & 0.0589 & 0.0611 & 0.0611
          & 0.1341 & 0.1136 & 0.1086 & 0.1119 & 0.1175 & 0.1212\\
    cGAN (Adler \& \"Oktem \cite{Adler:18}) 
          & 0.0698 & 0.0614 & 0.0623 & 0.0667 & 0.0704 & 0.0727
          & 0.1407 & 0.1262 & 0.1252 & 0.1291 & 0.1334 & 0.1361\\
    cGAN (Ohayon et al. \cite{Ohayon:ICCVW:21}) 
          & 0.0532 & 0.0536 & 0.0539 & 0.0540 & 0.0534 & 0.0540
          & 0.1128 & 0.1143 & 0.1151 & 0.1155 & 0.1157 & 0.1158\\
    cGAN (Ours) 
          & 0.0418 & \bf 0.0379 & 0.0421 & 0.0476 & 0.0516 & 0.0539
          & 0.0906 & \bf 0.0877 & 0.0965 & 0.1063 & 0.1135 & 0.1177\\
    \bottomrule
  \end{tabular}%
  }
\end{table*}

\newcommand{\MRIavgsampsamp}[7]{
    \begin{figure}[t]
    \def\tabwidcol{21mm} 
    \def\tabwidrow{21mm} 
    \def\figwid{0.18\columnwidth}
    \def\figheight{0.54\columnwidth}
    \hspace{10mm}
    \hspace{\figwid}
        {\sf \scriptsize
        \begin{tabular}{>{\centering}p{\tabwidrow}>{\centering}p{\tabwidrow}>{\centering}p{\tabwidrow}>{\centering}p{\tabwidrow}}
        cGAN (ours) & cGAN (Ohayon) & cGAN (Adler) & Langevin (Jalal)
        \end{tabular}}\\[-0.5mm]
    \rotatebox{90}{\sf \scriptsize
        \begin{tabular}{>{\centering}p{\tabwidcol}>{\centering}p{\tabwidcol}>{\centering}p{\tabwidcol}}
        E2E-VarNet & Truth & Truth
        \end{tabular}}%
    \begin{tikzpicture}
        \node[anchor=south west,inner sep=0] at (0,0) {\includegraphics[height=\figheight,trim=10 10 5 10,clip]{figures/mri/body_figs/body_mri_fig_R_#7_left_#1.png}};
        \draw [-latex, line width=1pt, yellow] (#2,\fpeval{#3 + #6}) -- (#4,\fpeval{#5 + #6});
        \draw [-latex, line width=1pt, yellow] (#2,#3) -- (#4,#5);
    \end{tikzpicture}
    \hspace{2mm}
    \rotatebox{90}{\sf \scriptsize
        \begin{tabular}{>{\centering}p{\tabwidcol}>{\centering}p{\tabwidcol}>{\centering}p{\tabwidcol}}
        Sample & Sample & Average (P=32)
        \end{tabular}}%
    \begin{tikzpicture}
        \node[anchor=south west,inner sep=0] at (0,0) {\includegraphics[height=\figheight,trim=10 10 5 10,clip]{figures/mri/body_figs/body_mri_fig_R_#7_right_#1.png}};
        \draw [-latex, line width=1pt, yellow] (#2,\fpeval{#3 + 2*#6}) -- (#4,\fpeval{#5 + 2*#6});
        \draw [-latex, line width=1pt, yellow] (#2,\fpeval{#3 + #6}) -- (#4,\fpeval{#5 + #6});
        \draw [-latex, line width=1pt, yellow] (#2,#3) -- (#4,#5);

        \draw [-latex, line width=1pt, yellow] (\fpeval{#2 + #6},\fpeval{#3 + 2*#6}) -- (\fpeval{#4 + #6},\fpeval{#5 + 2*#6});
        \draw [-latex, line width=1pt, yellow] (\fpeval{#2 + #6},\fpeval{#3 + #6}) -- (\fpeval{#4 + #6},\fpeval{#5 + #6});
        \draw [-latex, line width=1pt, yellow] (\fpeval{#2 + #6},#3) -- (\fpeval{#4 + #6},#5);

        \draw [-latex, line width=1pt, yellow] (\fpeval{#2 + 2*#6},\fpeval{#3 + 2*#6}) -- (\fpeval{#4 + 2*#6},\fpeval{#5 + 2*#6});
        \draw [-latex, line width=1pt, yellow] (\fpeval{#2 + 2*#6},\fpeval{#3 + #6}) -- (\fpeval{#4 + 2*#6},\fpeval{#5 + #6});
        \draw [-latex, line width=1pt, yellow] (\fpeval{#2 + 2*#6},#3) -- (\fpeval{#4 + 2*#6},#5);

        \draw [-latex, line width=1pt, yellow] (\fpeval{#2 + 3*#6},\fpeval{#3 + 2*#6}) -- (\fpeval{#4 + 3*#6},\fpeval{#5 + 2*#6});
        \draw [-latex, line width=1pt, yellow] (\fpeval{#2 + 3*#6},\fpeval{#3 + #6}) -- (\fpeval{#4 + 3*#6},\fpeval{#5 + #6});
        \draw [-latex, line width=1pt, yellow] (\fpeval{#2 + 3*#6},#3) -- (\fpeval{#4 + 3*#6},#5);
    \end{tikzpicture}
    \caption{Example $R=#7$ MRI reconstructions. Arrows show meaningful variations across samples.}
    \label{fig:avgsampsamp_#1}
    \end{figure}
}

\newcommand{\MRIavgerrstd}[2]{
    \begin{figure}[t]
    \def\tabwidcol{21mm} 
    \def\tabwidrow{19mm} 
    \def\figwid{0.18\columnwidth}
    \rotatebox{0}{\sf \scriptsize
        \begin{tabular}{>{\centering}p{\tabwidrow}>{\centering}p{\tabwidrow}>{\centering}p{\tabwidrow}>{\centering}p{\tabwidrow}>{\centering}p{\tabwidrow}>{\centering}p{\tabwidrow}}
        Truth & E2E-VarNet & cGAN (ours) & cGAN (Ohayon) & cGAN (Adler) & Langevin (Jalal)
        \end{tabular}
    }
    \\
    \includegraphics[width=\columnwidth,trim=10 10 10 10,clip]{figures/mri/body_figs/body_mri_fig_R_#2_avg_err_std_#1.png}
    \caption{Example $R=#2$ MRI reconstructions with $P=32$. 
    Row one: $P$-sample average $\hvec{x}\avgP$. 
    Row two: pixel-wise absolute error $|\hvec{x}\avgP-\vec{x}|$.
    Row three: pixel-wise SD $(\frac{1}{P}\sum_{i=1}^P(\hvec{x}_i-\hvec{x}\avgP)^2)^{1/2}$.}
    \label{fig:avgerrstd_#1}
    \end{figure}
}

\MRIavgsampsamp{19}{2.0}{1.65}{1.5}{1.5}{2.55}{8}
\MRIavgerrstd{19}{8}

\Figref{avgsampsamp_19} shows zoomed versions of two posterior samples $\hvec{x}_i$, as well as $\hvec{x}\avgP$, at $P=32$ and $R=8$.
The posterior samples show meaningful variations for the proposed method, essentially no variation for Ohayon et al.'s cGAN, and vertical or horizontal reconstruction artifacts for Adler \& \"Oktem's cGAN and the Langevin method, respectively.
The $\hvec{x}\avgP$ plots show that these artifacts are mostly suppressed by sample averaging with large $P$.

\Figref{avgerrstd_19} shows examples of $\hvec{x}\avgP$, along with the corresponding pixel-wise absolute errors $|\hvec{x}\avgP-\vec{x}|$ and pixel-wise SD images $(\frac{1}{P}\sum_{i=1}^P (\hvec{x}\avgP-\hvec{x}_i)^2)^{1/2}$, for $P=32$ and $R=8$.
The absolute-error image for the Langevin technique looks more diffuse than those of the other methods in the brain region.
The fact that it is brighter in the air region (i.e., near the edges) is a consequence of minor differences in sensitivity-map estimation.
The pixel-wise SD images show a lack of variability for the E2E-VarNet, which does not generate posterior samples, as well as Ohayon et al.'s cGAN, due to mode collapse.
The Langevin pixel-wise SD images show localized hot-spots that appear to be reconstruction artifacts. 

Appendices \ref{app:mriplotsR4} and \ref{app:mriplotsR8} show other example MRI recoveries with zoomed pixel-wise SD images at $R=4$ and $R=8$, respectively. 
Notably, Figures \ref{fig:mriapp_11} and \ref{fig:mriapp_9} show strong hallucinations for Langevin recovery at $R=8$, as highlighted by the red arrows.

\subsection{Inpainting experiments} \label{sec:inpainting}
In this section, our goal is to complete a large missing square in a face image.

\textbf{Data.~}
We used $256 \times 256$ CelebA-HQ face images \cite{Karras:ICLR:18} and a centered $128\times 128$ missing square.
We randomly split the dataset, yielding $27\,000$ training, $2\,000$ validation, and $1\,000$ testing images. 

\textbf{Architecture.~} 
For our cGAN, we use the CoModGAN generator and discriminator from \cite{Zhao:ICLR:21} with our proposed $\LonestdPt$ regularization. Unlike \cite{Zhao:ICLR:21}, we do not use MBSD \cite{Karras:ICLR:18} at the discriminator.

\begin{table*}[t]
  \caption{Average results for inpainting:
  FID was 
  computed from $1\,000$ test images with $P\!=\!32$,
  while CFID was computed from all 30\,000 images with $P\!=\!1$}
  \label{tab:inpainting256}
  \centering
  \vspace{0.1in}
  \resizebox{0.6\columnwidth}{!}{%
  \begin{tabular}{llll}
    \toprule
    Model & CFID$\downarrow$ & FID$\downarrow$ 
        & Time (128)$\downarrow$\\
    \midrule
    Score-SDE (Song et al.\ \cite{Song:ICLR:21}) & 5.11 & 7.92 
        & 48 min\\
    CoModGAN (Zhao et al.\ \cite{Zhao:ICLR:21}) & 5.29 & 8.50 
        & \textbf{217 ms}\\
    cGAN (ours) & \textbf{4.69} & \textbf{7.45} 
        & \textbf{217 ms}\\
    \bottomrule
  \end{tabular}%
  }
\end{table*}

\textbf{Training/validation/testing.~}
We use the same general training and testing procedure described in \secref{MRIexperiments}, but with $\betaadv=$ 5e-3, $n\batch=100$, and 110 epochs of cGAN training.
Running PyTorch on a server with 4 Tesla A100 GPUs, each with 82~GB of memory, the cGAN training takes approximately 2 days.
FID was evaluated on $1\,000$ test images using $P\!=\!32$ samples per measurement.
To avoid the bias that would result from evaluating CFID on only $1\,000$ images (see \appref{CFID}), CFID was evaluated on all $30\,000$ images with $P=1$. 

\textbf{Competitors.~} 
We compare with the state-of-the-art CoModGAN \cite{Zhao:ICLR:21}
and Score-based SDE \cite{Song:ICLR:21} approaches.
For CoModGAN, we use the PyTorch implementation from \cite{Zeng:github:22}. 
CoModGAN differs from our cGAN only in its use of MBSD and lack of $\LonestdPt$ regularization.
For Song et al.'s SDE, we use the authors' implementation from \cite{Song:github:21b} with their pretrained weights and the settings they suggested for the $256\times256$ CelebA-HQ dataset.

\textbf{Results.~} 
\tabref{inpainting256} shows test CFID, FID, and 128-sample generation time.
The table shows that our approach gave the best CFID and FID, and that the cGANs generated samples 13\,000 times faster than the score-based method.
\Figref{inpainting_22} shows an example of five generated samples for the three methods under test. 
The samples are all quite good, although a few generated by CoModGAN and the score-based technique have minor artifacts.  
Some samples generated by our technique show almond-shaped eyes, demonstrating fairness.
Additional examples are given in \appref{inpaintingplots}.

\newcommand{\inpaintingFigRow}[2]{
    \def\fighgt{21mm}
    \begin{figure}[#2]
    \centering
    {\scriptsize\sf
    \rotatebox{90}{\hspace{6mm} Original}%
        \includegraphics[height=\fighgt]{figures/inpainting/original_#1.png}
    \hspace{3mm}
    \rotatebox{90}{\hspace{1mm} \textcolor{white}{Sg}cGAN (ours)\textcolor{white}{N}}\,%
        \includegraphics[height=\fighgt,trim=10 9 10 6,clip]
            {figures/inpainting/5_recons_ours_#1.png}\\[0mm]
    \rotatebox{90}{\hspace{3mm} \textcolor{white}{Sg}Masked\textcolor{white}{N}}%
        \includegraphics[height=\fighgt]{figures/inpainting/masked_#1.png}
    \hspace{3mm}
    \rotatebox{90}{\hspace{0mm} \textcolor{white}{Sg}CoModGAN \textcolor{white}{N}}\,%
        \includegraphics[height=\fighgt,trim=10 9 10 6,clip]
            {figures/inpainting/5_recons_comodgan_#1.png}\\[0mm]
    \hspace{7mm}\hspace{\fighgt}
    \rotatebox{90}{\hspace{0mm} \textcolor{white}{Sg}Score-SDE\textcolor{white}{N}}\,%
        \includegraphics[height=\fighgt,trim=10 9 10 6,clip]
            {figures/inpainting/5_recons_langevin_#1.png}}
    \vspace{-2mm}
    \caption{Example of inpainting a $128\!\times\!128$ square on a $256\!\times\!256$ resolution CelebA-HQ image.}
    \label{fig:inpainting_#1}
\end{figure}
}

\inpaintingFigRow{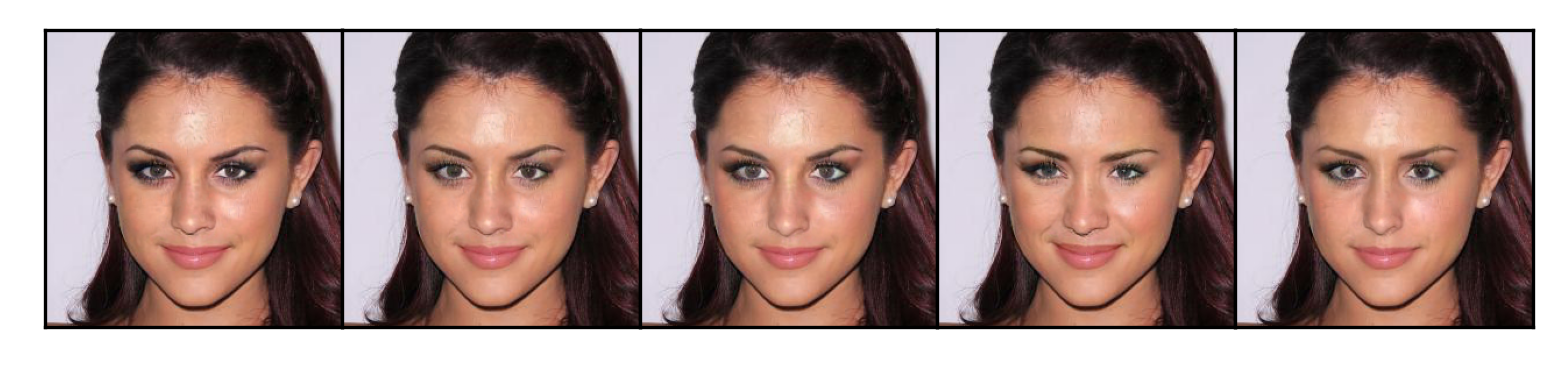}{t}

\section{Conclusion} \label{sec:conclusion}
We propose a novel regularization framework for image-recovery cGANs that consists of supervised-$\ell_1$ loss plus an appropriately weighted standard-deviation reward.
For the case of an independent Gaussian posterior, we proved that our regularization yields generated samples that agree with the true-posterior samples in both mean and covariance.
We also established limitations for alternatives based on supervised-$\ell_2$ regularization with or without a variance reward.
For practical datasets, we proposed a method to auto-tune our standard-deviation reward weight.

Experiments on parallel MRI and large-scale face inpainting showed our proposed method outperforming all cGAN and score-based competitors in CFID, which measures posterior-approximation quality, as well as other metrics like FID, PSNR, SSIM, LPIPS, and DISTS.
Furthermore, it generates samples thousands of times faster than Langevin/score-based approaches.

In ongoing work, we are extending our approach so that it can be trained to handle a wide range of recovery tasks, such as MRI with a generic acceleration and sampling mask \cite{Bendel:NIPSW:23}, or inpainting with a generic mask.
We are also extending our approach to other applications, such as super-resolution, deblurring, compressive sensing, denoising, and phase retrieval.

\textbf{Limitations.~} 
We acknowledge several limitations of our work.
First, while our current work focuses on how to build a fast and accurate posterior sampler, it's not yet clear how to best exploit the resulting posterior samples in each given application.
For example, in MRI, where the posterior distribution has the potential to help assess uncertainty in image recovery, it's still not quite clear how to best convey uncertainty information to radiologists (e.g., they may not gain much from pixel-wise SD images).
More work is needed on this front.
Second, we acknowledge that, because radiologists are risk-averse, more studies are needed before they will feel comfortable incorporating generative deep-learning-based methods into the clinical workflow.
Third, we acknowledge that the visual quality of our $R=8$ MRI reconstructions falls below clinical standards.
Fourth, some caution is needed when interpreting our CFID, FID, and DISTS perceptual metrics because the VGG-16 backbone used to compute them was trained on ImageNet data. 
Although there is some evidence that the resulting DISTS metric correlates well with radiologists' perceptions \cite{Kastryulin:22}, there is also evidence that ImageNet-trained features may discard information that is diagnostically relevant in medical imaging \cite{Kynkaanniemi:22}.
Thus our results need to be validated with a pathology-centric radiologist study before they can be considered relevant to clinical practice.

\begin{ack}
The authors are funded in part by the National Institutes of Health under grant R01-EB029957 and the National Science Foundation under grant CCF-1955587. 
\end{ack}

\bibliographystyle{ieeetr}
\small
\bibliography{macros_abbrev,books,misc,sparse,mri,machine}

\begin{thebibliography}{10}

\bibitem{Lehmann:Book:06}
E.~L. Lehmann and G.~Casella, {\em Theory of point estimation}.
\newblock Springer Science \& Business Media, 2006.

\bibitem{Blau:CVPR:18}
Y.~Blau and T.~Michaeli, ``The perception-distortion tradeoff,'' in {\em Proc.
  IEEE Conf. Comp. Vision Pattern Recog.}, pp.~6228--6237, 2018.

\bibitem{Sonderby:ICLR:17}
C.~K. S{\o}nderby, J.~Caballero, L.~Theis, W.~Shi, and F.~Husz{\'a}r,
  ``Amortised {MAP} inference for image super-resolution,'' in {\em Proc. Int.
  Conf. on Learn. Rep.}, 2017.

\bibitem{Mehrabi:CS:21}
N.~Mehrabi, F.~Morstatter, N.~Saxena, K.~Lerman, and A.~Galstyan, ``A survey on
  bias and fairness in machine learning,'' {\em ACM Comput. Surveys}, vol.~54,
  no.~6, pp.~1--35, 2021.

\bibitem{Sanchez:NIPSW:20}
T.~Sanchez, I.~Krawczuk, Z.~Sun, and V.~Cevher, ``Uncertainty-driven adaptive
  sampling via {GAN}s,'' in {\em Proc. Neural Inf. Process. Syst. Workshop},
  2020.

\bibitem{Chang:ICLR:19}
C.-H. Chang, E.~Creager, A.~Goldenberg, and D.~Duvenaud, ``Explaining image
  classifiers by counterfactual generation,'' in {\em Proc. Int. Conf. on
  Learn. Rep.}, 2019.

\bibitem{Isola:CVPR:17}
P.~Isola, J.-Y. Zhu, T.~Zhou, and A.~A. Efros, ``Image-to-image translation
  with conditional adversarial networks,'' in {\em Proc. IEEE Conf. Comp.
  Vision Pattern Recog.}, pp.~1125--1134, 2017.

\bibitem{Adler:18}
J.~Adler and O.~{\"O}ktem, ``Deep {B}ayesian inversion,'' {\em
  arXiv:1811.05910}, 2018.

\bibitem{Zhao:ICLR:21}
S.~Zhao, J.~Cui, Y.~Sheng, Y.~Dong, X.~Liang, E.~I.-C. Chang, and Y.~Xu,
  ``Large scale image completion via co-modulated generative adversarial
  networks,'' in {\em Proc. Int. Conf. on Learn. Rep.}, 2021.

\bibitem{Zhao:MIA:18}
H.~Zhao, H.~Li, S.~Maurer-Stroh, and L.~Cheng, ``Synthesizing retinal and
  neuronal images with generative adversarial nets,'' {\em Med. Image
  Analysis}, vol.~49, 07 2018.

\bibitem{Edupuganti:TMI:20}
V.~Edupuganti, M.~Mardani, S.~Vasanawala, and J.~Pauly, ``Uncertainty
  quantification in deep {MRI} reconstruction,'' {\em IEEE Trans. Med. Imag.},
  vol.~40, pp.~239--250, Jan. 2021.

\bibitem{Tonolini:JMLR:20}
F.~Tonolini, J.~Radford, A.~Turpin, D.~Faccio, and R.~Murray-Smith,
  ``Variational inference for computational imaging inverse problems,'' {\em J.
  Mach. Learn. Res.}, vol.~21, no.~179, pp.~1--46, 2020.

\bibitem{Sohn:NIPS:15}
K.~Sohn, H.~Lee, and X.~Yan, ``Learning structured output representation using
  deep conditional generative models,'' in {\em Proc. Neural Inf. Process.
  Syst. Conf.}, 2015.

\bibitem{Ardizzone:19}
L.~Ardizzone, C.~L{\"u}th, J.~Kruse, C.~Rother, and U.~K{\"o}the, ``Guided
  image generation with conditional invertible neural networks,'' {\em
  arXiv:1907.02392}, 2019.

\bibitem{Winkler:19}
C.~Winkler, D.~Worrall, E.~Hoogeboom, and M.~Welling, ``Learning likelihoods
  with conditional normalizing flows,'' {\em arXiv preprint arXiv:1912.00042},
  2019.

\bibitem{Sun:AAAI:21}
H.~Sun and K.~L. Bouman, ``Deep probabilistic imaging: {U}ncertainty
  quantification and multi-modal solution characterization for computational
  imaging,'' in {\em Proc. AAAI Conf. Artificial Intell.}, vol.~35,
  pp.~2628--2637, 2021.

\bibitem{Welling:ICML:11}
M.~Welling and Y.~W. Teh, ``Bayesian learning via stochastic gradient
  {L}angevin dynamics,'' in {\em Proc. Int. Conf. Mach. Learn.}, pp.~681--688,
  2011.

\bibitem{Song:NIPS:20}
Y.~Song and S.~Ermon, ``Improved techniques for training score-based generative
  models,'' in {\em Proc. Neural Inf. Process. Syst. Conf.}, 2020.

\bibitem{Jalal:NIPS:21}
A.~Jalal, M.~Arvinte, G.~Daras, E.~Price, A.~Dimakis, and J.~Tamir, ``Robust
  compressed sensing {MRI} with deep generative priors,'' in {\em Proc. Neural
  Inf. Process. Syst. Conf.}, 2021.

\bibitem{Song:ICLR:21}
Y.~Song, J.~Sohl-Dickstein, D.~P. Kingma, A.~Kumar, S.~Ermon, and B.~Poole,
  ``Score-based generative modeling through stochastic differential
  equations,'' in {\em Proc. Int. Conf. on Learn. Rep.}, 2021.

\bibitem{Song:ICLR:22}
Y.~Song, L.~Shen, L.~Xing, and S.~Ermon, ``Solving inverse problems in medical
  imaging with score-based generative models,'' in {\em Proc. Int. Conf. on
  Learn. Rep.}, 2022.

\bibitem{Soloveitchik:21}
M.~Soloveitchik, T.~Diskin, E.~Morin, and A.~Wiesel, ``Conditional {F}rechet
  inception distance,'' {\em arXiv:2103.11521}, 2021.

\bibitem{Ohayon:ICCVW:21}
G.~Ohayon, T.~Adrai, G.~Vaksman, M.~Elad, and P.~Milanfar, ``High perceptual
  quality image denoising with a posterior sampling { CGAN},'' in {\em Proc.
  IEEE Int. Conf. Comput. Vis. Workshops}, vol.~10, pp.~1805--1813, 2021.

\bibitem{Gulrajani:NIPS:17}
I.~Gulrajani, F.~Ahmed, M.~Arjovsky, V.~Dumoulin, and A.~Courville, ``Improved
  training of {W}asserstein {GAN}s,'' in {\em Proc. Neural Inf. Process. Syst.
  Conf.}, p.~5769–5779, 2017.

\bibitem{Mirza:14}
M.~Mirza and S.~Osindero, ``Conditional generative adversarial nets,'' {\em
  arXiv:1411.1784}, 2014.

\bibitem{Schonfeld:CVPR:20}
E.~Schonfeld, B.~Schiele, and A.~Khoreva, ``A {U-Net} based discriminator for
  generative adversarial networks,'' in {\em Proc. IEEE Conf. Comp. Vision
  Pattern Recog.}, pp.~8207--8216, 2020.

\bibitem{Karras:NIPS:20}
T.~Karras, M.~Aittala, J.~Hellsten, S.~Laine, J.~Lehtinen, and T.~Aila,
  ``Training generative adversarial networks with limited data,'' in {\em Proc.
  Neural Inf. Process. Syst. Conf.}, vol.~33, pp.~12104--12114, 2020.

\bibitem{Zhao:AAAI:21}
Z.~Zhao, S.~Singh, H.~Lee, Z.~Zhang, A.~Odena, and H.~Zhang, ``Improved
  consistency regularization for {GANs},'' in {\em Proc. AAAI Conf. Artificial
  Intell.}, vol.~35, pp.~11033--11041, 2021.

\bibitem{Karras:ICLR:18}
T.~Karras, T.~Aila, S.~Laine, and J.~Lehtinen, ``Progressive growing of {GAN}s
  for improved quality, stability, and variation,'' in {\em Proc. Int. Conf. on
  Learn. Rep.}, 2018.

\bibitem{Zhao:TCI:17}
H.~Zhao, O.~Gallo, I.~Frosio, and J.~Kautz, ``Loss functions for image
  restoration with neural networks,'' {\em IEEE Trans. Comput. Imag.}, vol.~3,
  pp.~47--57, Mar. 2017.

\bibitem{Ohayon:github:21}
G.~Ohayon, T.~Adrai, G.~Vaksman, M.~Elad, and P.~Milanfar, ``High perceptual
  quality image denoising with a posterior sampling {CGAN}.'' Downloaded from
  \url{https://github.com/theoad/pscgan}, July 2021.

\bibitem{Heusel:NIPS:17}
M.~Heusel, H.~Ramsauer, T.~Unterthiner, B.~Nessler, and S.~Hochreiter, ``{GAN}s
  trained by a two time-scale update rule converge to a local {N}ash
  equilibrium,'' in {\em Proc. Neural Inf. Process. Syst. Conf.}, vol.~30,
  2017.

\bibitem{Zbontar:18}
J.~Zbontar, F.~Knoll, A.~Sriram, M.~J. Muckley, M.~Bruno, A.~Defazio,
  M.~Parente, K.~J. Geras, J.~Katsnelson, H.~Chandarana, Z.~Zhang, M.~Drozdzal,
  A.~Romero, M.~Rabbat, P.~Vincent, J.~Pinkerton, D.~Wang, N.~Yakubova,
  E.~Owens, C.~L. Zitnick, M.~P. Recht, D.~K. Sodickson, and Y.~W. Lui,
  ``fast{MRI: An} open dataset and benchmarks for accelerated {MRI},'' {\em
  arXiv:1811.08839}, 2018.

\bibitem{Zhang:MRM:13}
T.~Zhang, J.~M. Pauly, S.~S. Vasanawala, and M.~Lustig, ``Coil compression for
  accelerated imaging with {C}artesian sampling,'' {\em Magn. Reson. Med.},
  vol.~69, no.~2, pp.~571--582, 2013.

\bibitem{Joshi:22}
M.~Joshi, A.~Pruitt, C.~Chen, Y.~Liu, and R.~Ahmad, ``Technical report
  (v1.0)--pseudo-random cartesian sampling for dynamic {MRI},'' {\em
  arXiv:2206.03630}, 2022.

\bibitem{Ronneberger:MICCAI:15}
O.~Ronneberger, P.~Fischer, and T.~Brox, ``{U-Net: C}onvolutional networks for
  biomedical image segmentation,'' in {\em Proc. Intl. Conf. Med. Image Comput.
  Comput. Assist. Intervent.}, pp.~234--241, 2015.

\bibitem{Sriram:MICCAI:20}
A.~Sriram, J.~Zbontar, T.~Murrell, A.~Defazio, C.~L. Zitnick, N.~Yakubova,
  F.~Knoll, and P.~Johnson, ``End-to-end variational networks for accelerated
  {MRI} reconstruction,'' in {\em Proc. Intl. Conf. Med. Image Comput. Comput.
  Assist. Intervent.}, pp.~64--73, 2020.

\bibitem{Jalal:github:21}
A.~Jalal, M.~Arvinte, G.~Daras, E.~Price, A.~Dimakis, and J.~Tamir,
  ``csgm-mri-langevin.'' \url{https://github.com/utcsilab/csgm-mri-langevin},
  2021.
\newblock Accessed: 2021-12-05.

\bibitem{Prussmann:MRM:99}
K.~P. Pruessmann, M.~Weiger, M.~B. Scheidegger, and P.~Boesiger, ``{SENSE}:
  {S}ensitivity encoding for fast {MRI},'' {\em Magn. Reson. Med.}, vol.~42,
  no.~5, pp.~952--962, 1999.

\bibitem{Uecker:MRM:14}
M.~Uecker, P.~Lai, M.~J. Murphy, P.~Virtue, M.~Elad, J.~M. Pauly, S.~S.
  Vasanawala, and M.~Lustig, ``{ESPIRiT}--an eigenvalue approach to
  autocalibrating parallel {MRI: W}here {SENSE} meets {GRAPPA},'' {\em Magn.
  Reson. Med.}, vol.~71, no.~3, pp.~990--1001, 2014.

\bibitem{Kastryulin:22}
S.~Kastryulin, J.~Zakirov, N.~Pezzotti, and D.~V. Dylov, ``Image quality
  assessment for magnetic resonance imaging,'' {\em arXiv:2203.07809}, 2022.

\bibitem{Zhang:CVPR:18}
R.~Zhang, P.~Isola, A.~A. Efros, E.~Shechtman, and O.~Wang, ``The unreasonable
  effectiveness of deep features as a perceptual metric,'' in {\em Proc. IEEE
  Conf. Comp. Vision Pattern Recog.}, pp.~586--595, 2018.

\bibitem{Ding:TPAMI:20}
K.~Ding, K.~Ma, S.~Wang, and E.~P. Simoncelli, ``Image quality assessment:
  {U}nifying structure and texture similarity,'' {\em IEEE Trans. Pattern Anal.
  Mach. Intell.}, vol.~44, no.~5, pp.~2567--2581, 2020.

\bibitem{Zeng:github:22}
Y.~Zeng, ``co-mod-gan-pytorch.'' Downloaded from
  \url{https://github.com/zengxianyu/co-mod-gan-pytorch}, Sept. 2022.

\bibitem{Song:github:21b}
Y.~Song, J.~Sohl-Dickstein, D.~P. Kingma, A.~Kumar, S.~Ermon, and B.~Poole,
  ``Score-based generative modeling through stochastic differential
  equations.'' Downloaded from
  \url{https://github.com/yang-song/score_sde_pytorch}, Oct. 2022.

\bibitem{Bendel:NIPSW:23}
M.~C. Bendel, R.~Ahmad, and P.~Schniter, ``Mask-agnostic posterior sampling
  {MRI} via conditional {GANs} with guided reconstruction,'' in {\em Proc.
  NeurIPS Workshop on Deep Inverse Problems}, 2023.

\bibitem{Kynkaanniemi:22}
T.~Kynk{\"a}{\"a}nniemi, T.~Karras, M.~Aittala, T.~Aila, and J.~Lehtinen, ``The
  role of imagenet classes in fr{\'e}chet inception distance,'' {\em
  arXiv:2203.06026}, 2022.

\bibitem{Leone:Techno:61}
F.~C. Leone, L.~S. Nelson, and R.~Nottingham, ``The folded normal
  distribution,'' {\em Technometrics}, vol.~3, no.~4, pp.~543--550, 1961.

\bibitem{Simonyan:14}
K.~Simonyan and A.~Zisserman, ``Very deep convolutional networks for
  large-scale image recognition,'' {\em arXiv:1409.1556}, 2014.

\bibitem{Szegedy:CVPR:16}
C.~Szegedy, V.~Vanhoucke, S.~Ioffe, J.~Shlens, and Z.~Wojna, ``Rethinking the
  inception architecture for computer vision,'' in {\em Proc. IEEE Conf. Comp.
  Vision Pattern Recog.}, 2016.

\bibitem{Ong:SigPy:19}
F.~Ong and M.~Lustig, ``{SigPy: A} python package for high performance
  iterative reconstruction,'' in {\em Proc. Annu. Meeting ISMRM}, vol.~4819,
  2019.

\bibitem{Chen:ECCV:20}
D.~Chen and M.~E. Davies, ``Deep decomposition learning for inverse imaging
  problems,'' in {\em Proc. European Conf. Comp. Vision}, pp.~510--526, 2020.

\bibitem{Kingma:ICLR:15}
D.~P. Kingma and J.~Ba, ``Adam: {A} method for stochastic optimization,'' in
  {\em Proc. Int. Conf. on Learn. Rep.}, 2015.

\bibitem{Deora:CVPRW:20}
P.~Deora, B.~Vasudeva, S.~Bhattacharya, and P.~M. Pradhan, ``Structure
  preserving compressive sensing {MRI} reconstruction using generative
  adversarial networks,'' in {\em Proc. IEEE Conf. Comp. Vision Pattern Recog.
  Workshop}, pp.~2211--2219, June 2020.

\end{thebibliography}
\normalsize

\clearpage
\appendix
\counterwithin{figure}{section}
\counterwithin{equation}{section}
\begin{center}
\Large \bf Supplementary Materials
\end{center}

\section{Posterior samples facilitate calibrated/fair detection} \label{app:detection}

Say that $s\in\{1,0\}$ denotes the presence or absence of a particular pathology (e.g., brain tumor), and say that we train a soft classifier $c(\cdot)$ on ground-truth images $\vec{x}$ and calibrate it such that
\begin{align}
c(\vec{x})
&= \Pr\{s=1|\vec{x}\} 
\label{eq:classifier}.
\end{align}
Now say that we observe a distorted/corrupted/incomplete measurement $\vec{y}=\mc{M}(\vec{x})$.
We would like to infer the probability that the pathology is present given $\vec{y}$, i.e., compute $\Pr\{s=1|\vec{y}\}$.
Note that 
\begin{align}
\Pr\{s=1|\vec{y}\}
&= \int \Pr\{s=1,\vec{x}|\vec{y}\} \deriv\vec{x}
= \int \Pr\{s=1|\vec{x},\vec{y}\} p(\vec{x}|\vec{y}) \deriv\vec{x} \\
&= \int \Pr\{s=1|\vec{x}\} p(\vec{x}|\vec{y}) \deriv\vec{x} 
= \int c(\vec{x}) p(\vec{x}|\vec{y}) \deriv\vec{x}
= \E\{c(\vec{x})|\vec{y}\} \\
&= \lim_{P\rightarrow\infty} \frac{1}{P}\sum_{i=1}^P c(\hvec{x}_i)
\text{~~for~~} \hvec{x}_i\sim\text{i.i.d.~} p(\vec{x}|\vec{y})
\label{eq:Ps|y},
\end{align}
where the last equality follows from the law of large numbers.
So, given access to many independent posterior samples $\{\hvec{x}_i\}$, equation \eqref{Ps|y} says that we can simply plug them into our calibrated classifier $c(\cdot)$ and average the result to compute $\Pr\{s=1|\vec{y}\}$.
Conversely, if we have access to only the posterior mean $\hvec{x}\mmse = \E\{\vec{x}|\vec{y}\}$, then because
\begin{align}
\Pr\{s=1|\vec{y}\}
= \E\{c(\vec{x})|\vec{y}\}
\neq c(\E\{\vec{x}|\vec{y}\})
= c(\hvec{x}\mmse)
\end{align}
for any non-linear $c(\cdot)$, the plug-in probability estimate will be incorrect.
In fact, there exists no point estimate $\hvec{x}$ that gives the correct $\Pr\{s=1|\vec{y}\}$ for general $c(\cdot)$.

Although above we defined $c(\cdot)$ as a (soft) binary pathology classifier, the same results hold if we define $c(\cdot)$ as a K-ary classifier of any protected attribute, such as race, gender, etc.
This implies that, if we have a machine-learning system that has been calibrated to classify fairly on clean ground-truth data $\vec{x}$, then the use of posterior samples $\{\hvec{x}_i\}$ enables it to classify fairly on distorted/corrupted/incomplete measurements $\vec{y}=\mc{M}(\vec{x})$, whereas the use of generic point-estimates $\hvec{x}$ does not.

\section{Proof of \propref{L1std}} \label{app:L1std}

Here we prove \propref{L1std}.
To begin, for an $N$-pixel image, we rewrite \eqref{LoneP}-\eqref{LstdP} as
\begin{align}
\LoneP(\vec{\theta})
&= \textstyle \sum_{j=1}^N \Ey\big\{ 
\ExzPgy\big\{ |x_j-\frac{1}{P}\sum_{i=1}^P \hat{x}_{ij}|~\big|\vec{y}\big\} 
\big\} 
\label{eq:LonePB}\\
\LstdP(\vec{\theta})
&= \textstyle \sum_{j=1}^N \Ey\big\{ 
\EzPgy\big\{ \frac{\gamma_P}{P}\sum_{i=1}^P |\hat{x}_{ij}-\frac{1}{P}\sum_{k=1}^P \hat{x}_{kj}| ~\big|\vec{y}\big\} 
\big\}
\label{eq:LstdPB} ,
\end{align}
where $x_j\defn[\vec{x}]_j$, $\hat{x}_{ij}\defn[\hvec{x}_i]_j$, and
\begin{align}\textstyle
\gamma_P \defn \sqrt{\frac{\pi P}{2(P-1)}}
\label{eq:gammaP} .
\end{align}
To simplify the notation in the sequel, we will consider an arbitrary fixed value of $j$ and use the abbreviations 
\begin{align}
x_j &\rightarrow X,
&\hat{x}_{ij} \rightarrow \hat{X}_i
\label{eq:abbrev} .
\end{align}
Recall that $\vec{x}$ and $\{\hvec{x}_i\}$ are mutually independent when conditioned on $\vec{y}$ because the code vectors $\{\vec{z}_i\}$ are generated independently of both $\vec{x}$ and $\vec{y}$.
In the context of \propref{L1std}, we also assume that the vector elements $x_j$ and $\hat{x}_{ij}$ are independent Gaussian when conditioned on $\vec{y}$.
This implies that we can make the notational shift
\begin{align}
\pxgy(x_j|\vec{y}) &\rightarrow \mc{N}(X;\mu_0,\sigma_0^2) ,
&\pxhgy(\hat{x}_{ij}|\vec{y}) \rightarrow \mc{N}(\hat{X}_i;\mu,\sigma^2)
\label{eq:gauss} ,
\end{align}
where $X$ and $\{\hat{X}_i\}$ are mutually independent.
With this simplified notation, we note that $[\hvec{x}\mmse]_j\rightarrow \mu_0$, and that mode collapse corresponds to $\sigma=0$.

Furthermore, if $\vec{\theta}$ can completely control $(\mu,\sigma)$, then \eqref{goal} can be rewritten as
\begin{align} 
({\mu}_*,{\sigma}_*)
= \arg\min_{\mu,\sigma} \big\{ \LoneP(\mu,\sigma) - \betastd \LstdP(\mu,\sigma) \big\}
~\Rightarrow~
\begin{cases}
{\mu}_*=\mu_0 \\
{\sigma}_*=\sigma_0
\end{cases}
\label{eq:goalB}
\end{align}
with
\begin{align}
\LoneP(\mu,\sigma)
&= \textstyle \ExxP\{|X-\frac{1}{P}\sum_{i=1}^P \hat{X}_i|\} 
\label{eq:LonePC}\\
\LstdP(\mu,\sigma)
&= \textstyle \ExP\{\frac{\gamma_P}{P} \sum_{i=1}^P |\hat{X}_i-\frac{1}{P}\sum_{k=1}^P \hat{X}_k|\} 
\label{eq:LstdPC} .
\end{align}
Although $\sigma_*$ must be positive, it turns out that we do not need to enforce this in the optimization \eqref{goalB} because it will arise naturally.

To further analyze \eqref{LonePC} and \eqref{LstdPC}, we define 
\begin{align}
\hat{\mu} &\defn \textstyle \frac{1}{P}\sum_{i=1}^P \hat{X}_i\\
\hat{\sigma} &\defn \textstyle \frac{\gamma_P}{P} \sum_{i=1}^P |\hat{X}_i - \hat{\mu}|
\label{eq:sigmahat} .
\end{align}
The quantity $\hat{\mu}$ can be recognized as the unbiased estimate of the mean $\mu$ of $\hat{X}_i$, and we now show that $\hat{\sigma}$ is an unbiased estimate of the SD $\sigma$ of $\hat{X}_i$ in the case that $\hat{X}_i$ is Gaussian.
To see this, first observe that the i.i.d.\ $\mc{N}(\mu,\sigma^2)$ property of $\{\hat{X}_i\}$ implies that $\hat{X}_i-\hat{\mu}=(1-\frac{1}{P})\hat{X}_i-\frac{1}{P}\sum_{k\neq i}\hat{X}_k$ is Gaussian with mean zero and variance $(1-\frac{1}{P})^2\sigma^2 + \frac{P-1}{P^2}\sigma^2 = \frac{P-1}{P}\sigma^2$.
The variable $|\hat{X}_i-\hat{\mu}|$ is thus half-normal distributed with mean $\sqrt{\frac{2(P-1)}{\pi P}\sigma^2}$ \cite{Leone:Techno:61}.
Because $\{\hat{X}_i\}$ are i.i.d., the variable $\frac{1}{P}\sum_{i=1}^P|\hat{X}_i-\hat{\mu}|$ has the same mean as $|\hat{X}_i-\hat{\mu}|$.
Finally, multiplying $\frac{1}{P}\sum_{i=1}^P|\hat{X}_i-\hat{\mu}|$ by $\gamma_P$ yields $\hat{\sigma}$ from \eqref{sigmahat}, and multiplying its mean using the expression for $\gamma_P$ from \eqref{gammaP} implies 
\begin{align}
\E\{\hat{\sigma}\} = \sigma ,
\end{align}
and so $\hat{\sigma}$ is an unbiased estimator of $\sigma$, the SD of $\hat{X}_i$.

With the above definitions of $\hat{\mu}$ and $\hat{\sigma}$, the optimization cost in \eqref{goalB} can be written as
\begin{align}
\LoneP(\mu,\sigma) - \betastd \LstdP(\mu,\sigma)
&= \ExxP\big\{|X-\hat{\mu}|\big\} - \betastd\ExP\big\{\hat{\sigma}\big\} \\
&= \ExxP\big\{|X-\hat{\mu}|\big\} - \betastd\sigma 
\label{eq:JA},
\end{align}
where in the last step we exploited the unbiased property of $\hat{\sigma}$. 
To proceed further, we note that the i.i.d.\ Gaussian property of $\{\hat{X}_i\}$ implies $\hat{\mu}\sim\mc{N}(\mu,\sigma^2/P)$, after which the mutual independence of $\{\hat{X}_i\}$ and $X$ yields
\begin{align}
X-\hat{\mu}
\sim \mc{N}(\mu_0-\mu,\sigma_0^2 + \sigma^2/P) 
\label{eq:prefolded}.
\end{align}
Taking the absolute value of a Gaussian random yields a folded-normal random variable \cite{Leone:Techno:61}.
Using the mean and variance in \eqref{prefolded}, the expressions in \cite{Leone:Techno:61} yield
\begin{align}
\ExxP\big\{ |X-\hat{\mu}| \big\} 
&= \sqrt{\frac{2(\sigma_0^2+\sigma^2/P)}{\pi}}
    \exp\Big(-\frac{(\mu_0 - \mu)^2}{2(\sigma_0^2+\sigma^2/P)}\Big)
\nonumber\\&\quad
   + (\mu_0 - \mu)\erf\Big(\frac{\mu_0 - \mu}{\sqrt{2(\sigma_0^2 + \sigma^2/P})}\Big) .
\end{align}
Thus the optimization cost \eqref{JA} can be written as
\begin{align}
J(\mu,\sigma) 
&= \sqrt{\frac{2(\sigma_0^2+\sigma^2/P)}{\pi}}
    \exp\Big(-\frac{(\mu-\mu_0)^2}{2(\sigma_0^2+\sigma^2/P)}\Big)
\nonumber\\&\quad
    + (\mu-\mu_0)\erf\Big(\frac{\mu-\mu_0}{\sqrt{2(\sigma_0^2 + \sigma^2/P})}\Big)
   - \betastd \sigma 
\label{eq:JB}.
\end{align}
Since $J(\cdot,\cdot)$ is convex, the minimizer $(\mu_*,\sigma_*)=\arg\min_{\mu,\sigma} J(\mu,\sigma)$ satisfies $\nabla J(\mu_*,\sigma_*)=(0,0)$.
To streamline the derivation, we define
\begin{align}
c &\defn \sqrt{2(\sigma_0^2 + \sigma^2/P)/\pi},
&s \defn \sqrt{\sigma_0^2 + \sigma^2/P}
\end{align}
so that 
\begin{align}
J(\mu,\sigma)
&= c \exp\Big(-\frac{(\mu-\mu_0)^2}{2s^2}\Big)
   + (\mu-\mu_0)\erf\Big(\frac{\mu-\mu_0}{\sqrt{2s^2}}\Big)
   - \betastd \sigma .
\end{align}
Because $c$ and $s$ are invariant to $\mu$, we get
\begin{align}
\frac{\partial J(\mu,\sigma)}{\partial \mu}
&= -c \exp\Big(\!-\!\frac{(\mu-\mu_0)^2}{2s^2}\Big) \frac{\mu-\mu_0}{s^2}
   + \erf\Big(\frac{\mu-\mu_0}{\sqrt{2s^2}}\Big) 
    + (\mu-\mu_0)\frac{2}{\sqrt{\pi}}\exp\Big(\!-\!\frac{(\mu-\mu_0)^2}{2s^2}\Big),
\end{align}
which equals zero if and only if $\mu=\mu_0$.
Thus we have determined that $\mu_*=\mu_0$.
Plugging $\mu_*=\mu_0$ into \eqref{JB}, we find
\begin{align}
J(\mu_*,\sigma)
&= \sqrt{2(\sigma_0^2+\sigma^2/P)/\pi} - \betastd \sigma 
\label{eq:JC}.
\end{align}
Taking the derivative with respect to $\sigma$, we get 
\begin{align}
\frac{\partial J(\mu_*,\sigma)}{\partial \sigma}
&= \sqrt{ \frac{2}{\pi P(P\sigma_0^2/\sigma^2+1)} } - \betastd \\
&= \sqrt{ \frac{2}{\pi P(P\sigma_0^2/\sigma^2+1)} } - \sqrt{\frac{2}{\pi P (P+1)}} ,
\end{align}
where in the last step we applied the value of $\betastd$ from \eqref{betastdG}.
It can now be seen that $\frac{\partial J(\mu_*,\sigma)}{\partial \sigma}=0$ if and only if $\sigma=\sigma_0$,
which implies that $\sigma_*=\sigma_0$.
Thus we have established \eqref{goalB}, which completes the proof of \propref{L1std}.

\section{Derivation of \propref{L2P}} \label{app:L2P}

Here we prove \propref{L2P}.
To start, we establish some notation and conditional-mean properties:
%
%
\begin{equation}
\begin{array}{r@{~}l@{\qquad}r@{~}l}
\hvec{x}\mmse &\defn \Exgy\{\vec{x}|\vec{y}\} \\
\vec{e}\mmse &\defn \vec{x}-\hvec{x}\mmse, & 
\vec{0} &= \Exgy\{\vec{e}\mmse|\vec{y}\} \\ 
\hvec{x}_i(\vec{\theta}) &\defn G_{\vec{\theta}}(\vec{z}_i,\vec{y}), &
\ovec{x}(\vec{\theta}) &\defn \Ezigy\{\hvec{x}_i(\vec{\theta})|\vec{y}\} \\ 
\hvec{x}\avgP(\vec{\theta}) &\defn \textstyle \frac{1}{P} \sum_{i=1}^P \hvec{x}_i(\vec{\theta}), & 
\ovec{x}(\vec{\theta}) &= \EzPgy\{\hvec{x}\avgP(\vec{\theta})|\vec{y}\} \\ 
\vec{d}_i(\vec{\theta})) &\defn \hvec{x}_i(\vec{\theta})-\ovec{x}(\vec{\theta}), &
\vec{0} &= \Ezigy\{\vec{d}_i(\vec{\theta})|\vec{y}\}~\forall\vec{\theta} \\ 
\vec{d}\avgP(\vec{\theta}) &\defn \textstyle \frac{1}{P} \sum_{i=1}^P \vec{d}_i(\vec{\theta}), &
\vec{0} &= \EzPgy\{\vec{d}\avgP(\vec{\theta})|\vec{y}\}~\forall\vec{\theta} \\ 
\end{array}
\label{eq:quantities}
\end{equation}
Our first step is to write \eqref{LtwoP} as 
\begin{align}
\LtwoP(\vec{\theta})
&= \Ey\big\{ \ExzPgy\{\|\vec{x}-\hvec{x}\avgP(\vec{\theta})\|_2^2 | \vec{y}\} \big\} 
\label{eq:LtwoPerr1} .
\end{align}
Leveraging the fact that $\hvec{x}\mmse$ and $\ovec{x}(\vec{\theta})$ are deterministic given $\vec{y}$, we write the inner term in \eqref{LtwoPerr1} as
\begin{align}
\lefteqn{ \ExzPgy\{\|\vec{x}-\hvec{x}\avgP(\vec{\theta})\|_2^2 | \vec{y}\} }\nonumber\\
&= \ExzPgy\{\|\hvec{x}\mmse+\vec{e}\mmse-\ovec{x}(\vec{\theta})-\vec{d}\avgP(\vec{\theta})\|_2^2 | \vec{y}\} \\
&= \ExzPgy\{\|\hvec{x}\mmse-\ovec{x}(\vec{\theta})\|_2^2 | \vec{y}\} 
\nonumber\\&\quad
    + 2\real\ExzPgy\{(\hvec{x}\mmse-\ovec{x}(\vec{\theta}))\herm(\vec{e}\mmse-\vec{d}\avgP(\vec{\theta}))|\vec{y}\}
\nonumber\\&\quad
    + \ExzPgy\{\|\vec{e}\mmse-\vec{d}\avgP(\vec{\theta})\|_2^2 | \vec{y}\} \\
&= \|\hvec{x}\mmse-\ovec{x}(\vec{\theta})\|_2^2 
    + 2\real\big[(\hvec{x}\mmse-\ovec{x}(\vec{\theta}))\herm 
        \underbrace{\ExzPgy\{(\vec{e}\mmse-\vec{d}\avgP(\vec{\theta}))|\vec{y}\}}_{\displaystyle =\vec{0}}\big]
\nonumber\\&\quad
    + \ExzPgy\{\|\vec{e}\mmse-\vec{d}\avgP(\vec{\theta})\|_2^2 | \vec{y}\} \label{eq:Exdiff2a}\\
&= 
    \|\hvec{x}\mmse-\Ezigy\{\hvec{x}_i(\vec{\theta})|\vec{y}\}\|_2^2 
    + \ExzPgy\{\|\vec{e}\mmse-\vec{d}\avgP(\vec{\theta})\|_2^2 | \vec{y}\}
\label{eq:Exdiff2}.
\end{align}
where in \eqref{Exdiff2a} we used the fact that $\vec{d}\avgP$ and $\vec{e}\mmse$ are both zero-mean when conditioned on $\vec{y}$.
We now leverage the fact that $\{\vec{z}_i\}$ are independent of $\vec{x}$ and $\vec{y}$ to write
\begin{align}
\lefteqn{ \ExzPgy\{\|\vec{e}\mmse-\vec{d}\avgP(\vec{\theta})\|_2^2 | \vec{y}\} }\nonumber\\
&= \ExzPgy\{\|\vec{e}\mmse\|_2^2 | \vec{y}\}
    + 2\real\ExzPgy\{\vec{e}\mmse\herm\vec{d}\avgP(\vec{\theta}) | \vec{y}\}
    + \ExzPgy\{\|\vec{d}\avgP(\vec{\theta})\|_2^2 | \vec{y}\} \\
&= \Exgy\{\|\vec{e}\mmse\|_2^2 | \vec{y}\}
    + 2\real\big[{\underbrace{\Exgy\{\vec{e}\mmse | \vec{y}\}}_{\displaystyle =\vec{0}}}\herm
    \underbrace{\EzPgy\{\vec{d}\avgP(\vec{\theta}) | \vec{y}\}}_{\displaystyle =\vec{0}}\big] + \EzPgy\{\|\vec{d}\avgP(\vec{\theta})\|_2^2 | \vec{y}\} 
\label{eq:Eediff2}.
\end{align}
Finally, we can leverage the fact that $\{\vec{z}_i\}$ are i.i.d.\ to write
\begin{align}
\EzPgy\{\|\vec{d}\avgP(\vec{\theta})\|_2^2 | \vec{y}\}
&= \textstyle \EzPgy\{\|\frac{1}{P}\sum_{i=1}^P\vec{d}_i(\vec{\theta})\|_2^2 | \vec{y}\} \\
&= \textstyle \frac{1}{P^2} \sum_{i=1}^P \Ezigy\{\|\vec{d}_i(\vec{\theta})\|_2^2 | \vec{y}\} \\
&= \tfrac{1}{P} \Ezigy\{\|\vec{d}_i(\vec{\theta})\|_2^2 | \vec{y}\} \text{~for any $i$} \\
&= \tfrac{1}{P} \Ezigy\{\tr[\vec{d}_i(\vec{\theta})\vec{d}_i(\vec{\theta})\herm] | \vec{y}\} \\
&= \tfrac{1}{P} \tr\big[ \Ezigy\{\vec{d}_i(\vec{\theta})\vec{d}_i(\vec{\theta})\herm\} | \vec{y}\} \big] \\
&= \tfrac{1}{P} \tr\big[ \Covzigy\{\hvec{x}_i(\vec{\theta}) | \vec{y}\} \big] 
\label{eq:EeP2} .
\end{align}
Combining \eqref{LtwoPerr1}, \eqref{Exdiff2}, \eqref{Eediff2}, and \eqref{EeP2}, we get the bias-variance decomposition
\begin{align}
\LtwoP(\vec{\theta})
&= \textstyle \Ey\Big\{ 
    \|\hvec{x}\mmse-\Ezigy\{\hvec{x}_i(\vec{\theta})|\vec{y}\}\|_2^2 
    + \tfrac{1}{P} \tr\big[ \Covzigy\{\hvec{x}_i(\vec{\theta}) | \vec{y}\} \big]
    + \Exgy\{\|\vec{e}\mmse\|_2^2 | \vec{y}\} 
    \Big\}
\label{eq:LtwoPerr} .
\end{align}
We now see that if $\vec{\theta}$ has complete control over the $\vec{y}$-conditional mean and covariance of $\hvec{x}_i(\vec{\theta})$, then minimizing \eqref{LtwoPerr} over $\vec{\theta}$ will cause
\begin{align}
\Ezigy\{\hvec{x}_i(\vec{\theta})|\vec{y}\} &= \hvec{x}\mmse \\
\Covzigy\{\hvec{x}_i(\vec{\theta}) | \vec{y}\} &= \vec{0} ,
\end{align}
which proves \propref{L2P}.


\section{Derivation of \eqref{LvarPerr}} \label{app:LvarPerr}

To show that the expression for $\LvarP$ in \eqref{LvarPerr} holds, we first rewrite \eqref{LvarP} as
\begin{align}
\LvarP(\vec{\theta})
&= \textstyle \frac{1}{P-1}\sum_{i=1}^P \Ey\{ \EzPgy\{\|\hvec{x}_i(\vec{\theta})-\hvec{x}\avgP(\vec{\theta})\|_2^2 | \vec{y}\} 
\label{eq:LvarPB}
\end{align}
where the definitions from \eqref{quantities} imply
\begin{align}
\lefteqn{
\EzPgy\{\|\hvec{x}_i(\vec{\theta})-\hvec{x}\avgP(\vec{\theta})\|_2^2 | \vec{y}\}
}\nonumber\\
&= \EzPgy\{\|\ovec{x}(\vec{\theta}) + \vec{d}_i(\vec{\theta}) - \vec{d}\avgP(\vec{\theta}) - \ovec{x}(\vec{\theta})\|_2^2 | \vec{y}\} \\
&= \textstyle \EzPgy\{\|\vec{d}_i(\vec{\theta}) - \frac{1}{P}\sum_{j=1}^P\vec{d}_j(\vec{\theta})\|_2^2 | \vec{y}\} \\
&= \textstyle \EzPgy\{\|(1-\frac{1}{P})\vec{d}_i(\vec{\theta}) - \frac{1}{P}\sum_{j\neq i}\vec{d}_j(\vec{\theta})\|_2^2 | \vec{y}\} \\
&= \textstyle (1-\frac{1}{P})^2 \Ezigy\{\|\vec{d}_i(\vec{\theta})\|_2^2 | \vec{y}\} 
        + \frac{P-1}{P^2}\Ezigy\{\|\vec{d}_i(\vec{\theta})\|_2^2 | \vec{y}\} 
\label{eq:ExdiffB}\\
&= \textstyle \frac{P-1}{P} \Ezigy\{\|\vec{d}_i(\vec{\theta})\|_2^2 | \vec{y}\} \text{~for any $i$} 
\label{eq:ExdiffC}.
\end{align}
For \eqref{ExdiffB}, we leveraged the zero-mean and i.i.d.\ nature of $\{\vec{d}_i(\vec{\theta})\}$ conditioned on $\vec{y}$.
By plugging \eqref{ExdiffC} into \eqref{LvarPB}, we get
\begin{align}
\LvarP(\vec{\theta})
&= \textstyle \frac{1}{P}\sum_{i=1}^P \Ey\{ \Ezigy\{\|\vec{d}_i(\vec{\theta})\|_2^2 | \vec{y}\} \}\\
&= \textstyle \Ey\{ \Ezigy\{\|\vec{d}_i(\vec{\theta})\|_2^2 | \vec{y}\} \} 
\text{~for any $i$}
\label{eq:LvarPerrB} \\
&= \Ey\{ \tr [ \Covzigy\{\hvec{x}_i(\vec{\theta}) | \vec{y}\} ] \}
\label{eq:LvarPerrC} ,
\end{align}
where \eqref{LvarPerrB} follows because $\{\vec{d}_i(\vec{\theta})\}$ are i.i.d.\ when conditioned on $\vec{y}$
and \eqref{LvarPerrC} follows from manipulations similar to those used for \eqref{EeP2}. 

\section{Proof of \propref{ratio}} \label{app:ratio}
Here we prove \propref{ratio}.
Recall from \eqref{quantities} that
$\hvec{x}\mmse \defn \E\{\vec{x}|\vec{y}\}$
and $\vec{e}\mmse \defn \vec{x} - \hvec{x}\mmse$.
To reduce clutter, we will abbreviate $\vec{e}\mmse$ by $\vec{e}$ in this appendix.
Also, for true-posterior samples $\hvec{x}_i\sim \pxgy(\cdot|\vec{y})$, we define 
\begin{align}
\hvec{e}_i &\defn \hvec{x}_i-\hvec{x}\mmse 
\label{eq:ei}.
\end{align}
Then using $\hvec{x}\avgP\defn\frac{1}{P}\sum_{i=1}^P\hvec{x}_i$ and from $\mc{E}_P$ from \eqref{Eratio}, we have
\begin{align}
\mc{E}_P
&= \E\{ \|\hvec{x}\avgP-\vec{x}\|^2 |\vec{y}\} \\
&= \textstyle \E\{ \|(\frac{1}{P}\sum_{i=1}^P\hvec{x}_i)-\vec{x}\|^2 |\vec{y}\} 
\\
&= \textstyle \E\{ \|\frac{1}{P}\sum_{i=1}^P(\hvec{x}_i-\vec{x})\|^2 |\vec{y}\} 
\\
&= \textstyle \frac{1}{P^2}\E\{ \|\sum_{i=1}^P(\hvec{x}_i-\hvec{x}\mmse+\hvec{x}\mmse-\vec{x})\|^2 |\vec{y}\} \\
&= \textstyle \frac{1}{P^2}\E\{ \|\sum_{i=1}^P(\hvec{e}_i-\vec{e})\|^2 |\vec{y}\} \\
&= \textstyle \frac{1}{P^2}\E\{ \sum_{i=1}^P(\hvec{e}_i-\vec{e})\herm 
    \sum_{j=1}^P(\hvec{e}_j-\vec{e})|\vec{y}\} \\
&= \textstyle \frac{1}{P^2}\sum_{i=1}^P \sum_{j=1}^P \E\{(\hvec{e}_i-\vec{e})\herm 
    (\hvec{e}_j-\vec{e})|\vec{y}\} \\
&= \textstyle \frac{1}{P^2}\sum_{i=1}^P \E\{(\hvec{e}_i-\vec{e})\herm 
    (\hvec{e}_i-\vec{e})|\vec{y}\} 
    + \frac{1}{P^2}\sum_{i=1}^P \sum_{j\neq i} \E\{(\hvec{e}_i-\vec{e})\herm 
    (\hvec{e}_j-\vec{e})|\vec{y}\} \\
&= \textstyle \frac{1}{P^2}\sum_{i=1}^P \big[ 
    \E\{\|\hvec{e}_i\|^2|\vec{y}\}
    -2\real\E\{\hvec{e}_i\herm\vec{e}|\vec{y}\}
    +\E\{\|\vec{e}\|^2|\vec{y}\} \big]
    \nonumber\\&\quad\textstyle
    + \frac{1}{P^2}\sum_{i=1}^P \sum_{j\neq i} \real\big[
    \E\{\hvec{e}_i\herm\hvec{e}_j|\vec{y}\} 
    - \E\{\hvec{e}_i\herm\vec{e}|\vec{y}\} 
    - \E\{\vec{e}\herm\hvec{e}_j|\vec{y}\} 
    + \E\{\|\vec{e}\|^2|\vec{y}\} \big] \\
&= \textstyle \frac{1}{P^2}\sum_{i=1}^P 
    \E\{\|\hvec{e}_i\|^2|\vec{y}\}
    +\frac{1}{P}\E\{\|\vec{e}\|^2|\vec{y}\} 
    + \frac{P(P-1)}{P^2} \E\{\|\vec{e}\|^2|\vec{y}\} 
\label{eq:EP1},
\end{align}
where certain terms vanished because the i.i.d.\ and zero-mean properties of $\{\vec{e},\hvec{e}_1,\dots,\hvec{e}_P\}$ imply
\begin{align}
\E\{\hvec{e}_i\herm\hvec{e}_j|\vec{y}\}
&=\E\{\hvec{e}_i|\vec{y}\}\herm\E\{\hvec{e}_j|\vec{y}\} = 0\\
\E\{\hvec{e}_i\herm\vec{e}|\vec{y}\}
&=\E\{\hvec{e}_i|\vec{y}\}\herm\E\{\vec{e}|\vec{y}\} = 0\\
\E\{\vec{e}\herm\hvec{e}_j|\vec{y}\}
&=\E\{\vec{e}|\vec{y}\}\herm\E\{\hvec{e}_j|\vec{y}\} = 0 .
\end{align}
Finally, note that
$\E\{\|\vec{e}\|^2|\vec{y}\}=\mc{E}\mmse$ from \eqref{quantities}.
Furthermore, because $\{\vec{x},\hvec{x}_1,\dots,\hvec{x}_P\}$ are independent samples of $\pxgy(\cdot|\vec{y})$ under the assumptions of \propref{ratio}, we have $\E\{\|\vec{e}\|^2|\vec{y}\}=\E\{\|\hvec{e}_i\|^2|\vec{y}\}$ and so \eqref{EP1} becomes
\begin{align}
\mc{E}_P 
&= \frac{1}{P^2}\sum_{i=1}^P \mc{E}\mmse
    +\frac{1}{P} \mc{E}\mmse 
    + \frac{P(P-1)}{P^2} \mc{E}\mmse 
= \frac{P+1}{P} \mc{E}\mmse .
\end{align}
This result holds for any $P\geq 1$, which implies the ratio
\begin{align}
\frac{\mc{E}_1}{\mc{E}_P} = \frac{2P}{P+1} .
\end{align}

\section{CFID implementation details} \label{app:cfid}
\setcounter{table}{0}
\renewcommand\thetable{\thesection.\arabic{table}}

With the Gaussian approximation described in \secref{cfid}, where $\pxugyu$ and $\pxuhgyu$ are approximated by $\mc{N}(\muxugyu,\Sigxugyu)$ and $\mc{N}(\muxhugyu,\Sigxuhgyu)$, respectively, the CWD in \eqref{CWD} reduces to
\begin{align}
\CFID &\defn \Ey\big\{ \|\muxugyu-\muxhugyu\|_2^2  + \tr\big[ \Sigxugyu + \Sigxuhgyu - 2 \big( \Sigxugyu^{1/2} \Sigxuhgyu \Sigxugyu^{1/2} \big)^{1/2} \big]\big\}
\label{eq:CFID} .
\end{align}
The values in \eqref{CFID} are computed using 
\begin{align}
\muxugyu 
&= \muxu + \Sigxuyu\Sigyuyu^{-1}(\uvec{y}-\muyu)
\label{eq:muxugyu}\\
\Sigxugyu 
&= \Sigxuxu - \Sigxuyu\Sigyuyu^{-1}\Sigxuyu\tran
\label{eq:sigxxgy}\\
\muxhugyu 
&= \muxhu + \Sigxhuyu\Sigyuyu^{-1}(\uvec{y}-\muyu)\\
\Sigxuhgyu 
&= \Sigxhuxhu - \Sigxhuyu\Sigyuyu^{-1}\Sigxhuyu\tran
\label{eq:sigxuxugy}.
\end{align}
Plugging \eqref{muxugyu}-\eqref{sigxuxugy} into \eqref{CFID}, the CFID can be written as \cite[Lemma 2]{Soloveitchik:21}:
\begin{align}
\CFID 
&= \|\muxu-\muxhu\|_2^2 
+ \tr\Big[(\Sigxuyu - \Sigxhuyu)\Sigyuyu^{-1}(\Sigxuyu - \Sigxhuyu)\tran\Big] 
\nonumber\\&\quad
+ \tr\Big[\Sigxugyu + \Sigxuhgyu - 2 \big( \Sigxugyu^{1/2} \Sigxuhgyu \Sigxugyu^{1/2} \big)^{1/2}\Big] ,
\label{eq:CFID_2}
\end{align}
where $\Sigyuyu^{-1}$ is typically implemented using a pseudo-inverse.

We now detail how the means and covariances in \eqref{CFID_2} are computed.
We start with a dataset $\{(\vec{x}_t, \vec{y}_t)\}_{t=1}^n$ of truth/measurement pairs.
For each $\vec{y}_t$, we generate a set of $P$ posterior samples $\{\hvec{x}_{ti}\}_{i=1}^{P}$. 
We merge these samples with $P$ repetitions of $\vec{x}_t$ and $\vec{y}_t$ to obtain $\{(\vec{x}_{ti}, \vec{y}_{ti}, \hvec{x}_{ti})\}_{i=1}^{P}$ for $t=1\dots n$.
These terms are processed by a feature-generating network to yield the feature embeddings $\{(\uvec{x}_{ti}, \uvec{y}_{ti}, \huvec{x}_{ti})\}_{i=1}^{P}$, which are then packed into matrices $\uvec{X}$, $\uvec{Y}$, and $\huvec{X}$ with $P n$ rows.
We used the VGG-16 feature-generating network \cite{Simonyan:14} for our MRI experiments, since \cite{Kastryulin:22} found that it gave results that correlated much better with radiologists' perceptions, while we used the standard Inception-v3 network \cite{Szegedy:CVPR:16} for our inpainting experiments. 
The embeddings are then used to compute the sample-mean values
\begin{align} \textstyle
\muxu \defn \frac{1}{P n} \vec{1}\tran \uvec{X},
\quad
\muyu \defn \frac{1}{P n} \vec{1}\tran \uvec{Y},
\quad
\muxhu \defn \frac{1}{P n} \vec{1}\tran \huvec{X}
\label{eq:meancfid}.
\end{align}
We then subtract the sample mean from each row of $\uvec{X}$, $\uvec{Y}$, and $\huvec{X}$ to give the zero-mean embedding matrices $\uvec{X}\zm\defn\uvec{X}-\vec{1}\muxu\tran$, $\uvec{Y}\zm\defn\uvec{Y}-\vec{1}\muyu\tran$, and $\huvec{X}\zm\defn\huvec{X}-\vec{1}\muxhu\tran$, which are then used to compute the sample covariance matrices
\begin{subequations} \label{eq:covcfid}
\begin{align}
\Sigxuxu 
&\defn \textstyle \frac{1}{P n} \uvec{X}\zm\tran\uvec{X}\zm, \quad
\Sigyuyu
\defn \frac{1}{P n} \uvec{Y}\zm\tran\uvec{Y}\zm, \quad
\Sigxhuxhu
\defn \frac{1}{P n} \huvec{X}\zm\tran\huvec{X}\zm \\
\Sigxuyu
&\defn \textstyle \frac{1}{P n} \uvec{X}\zm\tran\uvec{Y}\zm , \quad
\Sigxhuyu
\defn \frac{1}{P n} \huvec{X}\zm\tran\uvec{Y}\zm .
\end{align}
\end{subequations}
We plug the sample statistics from \eqref{meancfid}-\eqref{covcfid} into \eqref{muxugyu}-\eqref{sigxuxugy}, which yields the statistics needed to compute the CFID in \eqref{CFID_2}. 
In \cite{Soloveitchik:21}, the authors use $P=1$ in all of their experiments. 
To be consistent with how we evaluated the other metrics, we use $P=32$ unless otherwise noted.


\section{MR imaging details} \label{app:MRI}
We now give details on magnetic resonance (MR) image recovery.
Suppose that the goal is to recover the $N$-pixel MR image $\vec{i}\in\Complex^N$ from the multicoil measurements $\{\vec{k}_c\}_{c=1}^C$, where \cite{Prussmann:MRM:99}
\begin{equation} 
\vec{k}_c = \vec{MFS}_c\vec{i} + \vec{n}_c
\label{eq:yc}.
\end{equation}
In \eqref{yc}, $C$ refers to the number of coils, $\vec{k}_c\in\Complex^M$ are the measurements from the $c$th coil, $\vec{M} \in \Real^{M\times N}$ is a sub-sampling operator containing rows from $\vec{I}_N$---the $N\times N$ identity matrix, $\vec{F} \in \Complex^{N\times N}$ is the unitary 2D discrete Fourier transform, $\vec{S}_c \in \Complex^{N\times N}$ is a diagonal matrix containing the sensitivity map of the $c$th coil,  and $\vec{n}_c \in \Complex^{M}$ is noise.
From \eqref{yc}, it can be seen that the MR measurements are collected in the spatial Fourier domain, otherwise known as the ``k-space.''
The sensitivity maps $\{\vec{S}_c\}$ are estimated from $\{\vec{k}_c\}$ using ESPIRiT \cite{Uecker:MRM:14} (in our case via SigPy \cite{Ong:SigPy:19}), which yields maps with the property $\sum_{c=1}^C\vec{S}_c\herm\vec{S}_c=\vec{I}_N$.
The ratio $R\triangleq \frac{N}{M}$ is known as the acceleration rate.

There are different ways that one could apply the generative posterior sampling framework to multicoil MR image recovery.
One is to configure the generator to produce posterior samples $\hvec{i}$ of the complex image $\vec{i}$. Another is to configure the generator to produce posterior samples $\hvec{x}$ of the stack $\vec{x}\defn [\vec{x}_1\tran,\dots,\vec{x}_C\tran]\tran$ of ``coil images'' $\vec{x}_c\defn\vec{S}_c\vec{i}$ and later coil-combining them to yield a complex image estimate $\hvec{i}\defn [\vec{S}\herm_1,\dots,\vec{S}\herm_C]\hvec{x}$.
We take the latter approach.
Furthermore, rather than feeding our generator with k-space measurements $\vec{k}_c$, we choose to feed it with aliased coil images $\vec{y}_c\defn \vec{F}\herm\vec{M}\tran\vec{k}_c$.
Writing \eqref{yc} in terms of the coil images, we obtain
\begin{equation} 
\vec{y}_c = \vec{F}\herm\vec{M}\tran\vec{MFx}_c + \vec{w}_c
\label{eq:yc2},
\end{equation}
where $\vec{w}_c\defn \vec{F}\herm\vec{M}\tran\vec{n}_c$.
Then we can stack $\{\vec{y}_c\}$ and $\{\vec{w}_c\}$ column-wise into vectors $\vec{y}$ and $\vec{w}$, and set $\vec{A} = \vec{I}_C \otimes \vec{F}\herm\vec{M}\tran\vec{MF}\in\Complex^{NC\times NC}$, to obtain the formulation $\vec{y} = \vec{Ax}+\vec{w}$ described in \secref{intro}.

To train our generator, we assume to have access to paired training examples $\{(\vec{x}_t,\vec{y}_t)\}$, where $\vec{x}_t$ is a stack of coil images and $\vec{y}_t$ is the corresponding stack of k-space coil measurements. 
The fastMRI multicoil dataset \cite{Zbontar:18} provides $\{(\vec{x}_t,\vec{k}_t)\}$, from which we can easily obtain $\{(\vec{x}_t,\vec{y}_t)\}$. 

\section{Data-consistency} \label{app:consistency}
In this section, we describe a data-consistency procedure that can be optionally used when our cGAN is used to solve a \emph{linear} inverse problem, i.e., to recover $\vec{x}$ from $\vec{y}$ under a model of the form
\begin{equation}
\vec{y}
= \vec{Ax}+\vec{w}
\label{eq:y} ,
\end{equation}
where $\vec{A}$ is a known linear operator and $\vec{w}$ is unknown noise.
The motivation is that, in some applications, such as medical imaging or inpainting, the end user may feel comfortable knowing that the generated samples $\{\hvec{x}_i\}$ are consistent with the measurements $\vec{y}$ in that 
\begin{align}
\vec{y}=\vec{A}\hvec{x}_i
\label{eq:DC}.
\end{align}
When \eqref{DC} holds, $\vec{A}^+\vec{y}=\vec{A}^+\vec{A}\hvec{x}_i$ must also hold, where $(\cdot)^+$ denotes the pseudo-inverse.
The quantity $\vec{A}^+\vec{A}$ can be recognized as the orthogonal projection matrix associated with the row space of $\vec{A}$.
So, \eqref{DC} requires the component of $\hvec{x}_i$ in the row space of $\vec{A}$ to equal $\vec{A}^+\vec{y}$, while placing no constraints on the component of $\hvec{x}_i$ in the nullspace of $\vec{A}$.
This suggests the following data-consistency procedure:
\begin{align}
\hvec{x}_i
&= (\vec{I}-\vec{A}^+\vec{A})\hvec{x}_i\raw + \vec{A}^+\vec{y}
\label{eq:DC2}.
\end{align}
where $\hvec{x}_i\raw$ is the raw generator output. 
We note that a version of this idea for point estimation was proposed in \cite{Sonderby:ICLR:17}.

The data-consistency procedure \eqref{DC2} ensures that any generative method will generate only the component of $\vec{x}$ that lies in the nullspace of $\vec{A}$. 
Consequently, \eqref{DC2} is admissible only when $\vec{A}$ has a non-trivial nullspace.
Also, because no attempt is made to remove the noise $\vec{w}$ in $\vec{y}$, this approach is recommended only for low-noise applications.
For high-noise applications, an extension based on the dual-decomposition approach \cite{Chen:ECCV:20} would be more appropriate, but we leave this to future work.

When applying \eqref{DC2} to the MRI formulation in \appref{MRI}, we note that $\vec{A}=\vec{I}_C \otimes \vec{F}\herm\vec{M}\tran\vec{MF}$ is an orthogonal projection matrix, and so $\vec{I}-\vec{A}^+\vec{A} = \vec{I}-\vec{A} = \vec{I}\otimes \vec{F}\herm(\vec{I}-\vec{M}\tran\vec{M})\vec{F}$.

\section{Implementation details} \label{app:implementation}
The code for our model can be found here: \url{https://github.com/matt-bendel/rcGAN}.

\subsection{MRI} \label{app:mriarchitecture}

\subsubsection{cGAN training} 

At each training iteration, our cGAN's generator takes in $n\batch$ measurement samples $\vec{y}_t$ and $P\train$ code vectors for every $\vec{y}_t$, and performs an optimization step on the loss
\begin{equation}
\Ell_{\mathsf{G}}(\vec{\theta}) 
\defn \betaadv \Ladv(\vec{\theta},\vec{\phi}) + \LonePt(\vec{\theta}) - \betastd\LstdPt(\vec{\theta})
\label{eq:genloss} ,
\end{equation}
where by default we use $\betaadv = 1\text{e-5}$, $n\batch=36$, $P\train=2$, and update $\betastd$ via \eqref{betastd} using $P\val=8$.
Then, using the $P\train n\batch$ generator outputs, our cGAN's discriminator performs an optimization step on the loss
\begin{equation}
\Ell_{\mathsf{D}}(\vec{\phi}) = -\Ladv(\vec{\theta},\vec{\phi}) + \alpha_1 \Lgp(\vec{\phi}) + \alpha_2 \Ldrift(\vec{\phi})
\label{eq:dloss} ,
\end{equation}
with gradient penalty $\Ell_\text{gp}$ from \cite{Gulrajani:NIPS:17}. 
As per \cite{Karras:ICLR:18}, $\Ell_\text{drift}$ is a drift penalty, $\alpha_1 = 10$, $\alpha_2 = 0.001$, and one discriminator update was used per generator update.
The models were trained for 100 epochs using the Adam optimizer \cite{Kingma:ICLR:15} with a learning rate of 1e-3, $\beta_1=0$, and $\beta_2=0.99$, as in \cite{Karras:ICLR:18}.
Running PyTorch on a server with 4 Tesla A100 GPUs, each with 82~GB of memory, the training of an MRI cGAN took approximately 1 day.

Adler and \"Oktem's cGAN \cite{Adler:18} uses generator loss $\betaadv\Ladler(\vec{\theta},\vec{\phi})$, where $\Ladler(\vec{\theta},\vec{\phi})$ was described in \eqref{Ladler}, and discriminator loss $-\Ladler(\vec{\theta},\vec{\phi})+ \alpha_1 \Lgp(\vec{\phi}) + \alpha_2 \Ldrift(\vec{\phi})$ 
with the values of $\alpha_1=$10, $\alpha_2=$0.001, and $\betaadv=$1, as in the original paper. 

Ohayon et al.'s cGAN \cite{Ohayon:ICCVW:21} uses generator loss $\betaadv\Ladv(\vec{\theta},\vec{\phi}) + \LtwoPt(\vec{\theta})$, where $\LtwoPt(\vec{\theta})$ was described in \eqref{LtwoP}, and discriminator loss $-\Ladv(\vec{\theta},\vec{\phi})+ \alpha_1 \Lgp(\vec{\phi}) + \alpha_2 \Ldrift(\vec{\phi})$
with the values $\alpha_1=$10, $\alpha_2=$0.001, and $\betaadv=$1e-5. We modify $\betaadv$ to re-balance the loss due to an increased magnitude of our discriminator's outputs.

All three cGANs used the same generator and discriminator architectures (detailed below), except that Adler and \"Oktem's discriminator used extra input channels to facilitate the 3-input loss $\Ladler(\vec{\theta},\vec{\phi})$ from \eqref{Ladler}.

\subsubsection{cGAN generator architecture} 
For our MRI experiments, we take inspiration from the UNet architecture from \cite{Ronneberger:MICCAI:15}, using it as the basis for the cGAN generators. 
The primary input $\vec{y}$ is concatenated with the code vector $\vec{z}$ and fed through the UNet. 
The network consists of 4 pooling layers with 128 initial channels.
However, instead of pooling, we opt to use convolutions with kernels of size $3\times3$, ``same'' padding, and a stride of 2 when downsampling. 
Likewise, we upsample using transpose convolutions, again with kernels of size $3\times3$, ``same'' padding, and a stride of 2.
All other convolutions utilize kernels of size $3\times3$, ``same'' padding, and a stride of 1.

Within each encoder and decoder layer we include a residual block, the architecture of which can be found in \cite{Adler:18}.
We use instance-norm for all normalization layers and parametric ReLUs as our activation functions, in which the network learns the optimal ``negative slope.''
Finally, we include 5 residual blocks at the base of the UNet, in between the encoder and decoder. This is done in an effort to artificially increase the depth of the network and is inspired by \cite{Deora:CVPRW:20}.
Our generator has $86\,734\,334$ trainable parameters.

\subsubsection{cGAN discriminator architecture} 
Our discriminator is a standard CNN with 6 layers and 1 fully-connected layer. 
In the first 3 layers, we use convolutions with kernels of size $3\times3$, ``same'' padding. We reduce spatial resolution with average pooling, using $2\times2$ kernels with a stride of 2. 
We use batch-norm as our normalization layer and leaky ReLUs with a ``negative-slope'' of 0.2 as our activation functions. The network outputs an estimated Wasserstein score for the whole image.


\subsubsection{E2E-VarNet}

For the Sriram et al.'s E2E-VarNet \cite{Sriram:MICCAI:20}, we use the same training procedure and hyperparameters outlined in \cite{Jalal:NIPS:21} other than replacing the sampling pattern with the GRO undersampling mask.
As in \cite{Jalal:NIPS:21}, we use the SENSE-based coil-combined image as the ground truth instead of the RSS image.



\subsubsection{Langevin approach} 
For Jalal et al.'s MRI approach \cite{Jalal:NIPS:21}, we do not modify the original implementation from \cite{Jalal:github:21} other than replacing the default sampling pattern with the GRO undersampling mask. 
We generated 32 samples for 72 different test images using a batch-size of 4, which took roughly 6 days. 
These samples were generated on a server with 4 NVIDIA V100 GPUs, each with 32~GB of memory.
We used 4 samples per batch (and recorded the time to generate 4 samples in \tabref{mriresults}) because the code from \cite{Jalal:github:21} is written to generate one sample per GPU.

\subsection{Inpainting} \label{app:inpainting}

\subsubsection{Our cGAN} 
For our generator and discriminator, we use the CoModGAN networks from \cite{Zhao:ICLR:21}.
Unlike CoModGAN, however, we train our cGAN with $\LonestdPt$ regularization and we do not use MBSD at the discriminator.
We use the same general training and testing procedure described in \secref{MRIexperiments}, but with $\betaadv=$ 5e-3, $n\batch=100$, and 110 epochs of cGAN training.
Running PyTorch on a server with 4 Tesla A100 GPUs, each with 82~GB of memory, the training takes approximately 2 days.

\subsubsection{CoModGAN} 
We use the PyTorch implementation of CoModGAN from \cite{Zeng:github:22} and train the model
to inpaint a $128\times128$ centered square on $256\times256$ CelebA-HQ images. 
The total training time on a server with 4 NVIDIA A100 GPUs, each with 82~GB of memory, is roughly 2 days.

\subsubsection{Score-based SDE} 
For the inpainting experiment in \secref{inpainting}, we compare against Song et al.'s more recent SDE technique \cite{Song:ICLR:21}, for which we use the publicly available pretrained weights, the suggested settings for the $256\times256$ CelebA-HQ dataset, and the code from the official PyTorch implementation \cite{Song:github:21b}.
We generate 32 samples for all $1\,000$ images in our test set, using a batch-size of 20 and generating 32 samples for each batch element concurrently.
The total generation time on a server with 4 NVIDIA A100 GPUs, each with 82~GB of memory, is roughly 9 days.


\section{Additional experimental results} \label{app:additionalresults}

\subsection{CFID decomposition into mean and covariance components} \label{app:CFID}
In this section, we investigate the small-sample bias effects of CFID, which have been previously noted in \cite{Soloveitchik:21}.
To do this, we write the CFID from \eqref{CFID} as a sum of two terms: a term that quantifies the conditional-mean error and a term that quantifies the conditional-covariance error:
\begin{align}
\CFID 
&= \CFID\mean + \CFID\cova \\
\CFID\mean
&\defn \Ey\{ \|\muxugyu-\muxhugyu\|_2^2 \} \\
\CFID\cova
&\defn  \tr\big[ \Sigxugyu + \Sigxuhgyu 
- 2 \big( \Sigxugyu^{1/2} \Sigxuhgyu \Sigxugyu^{1/2} \big)^{1/2} \big]
\label{eq:CFIDcov}.
\end{align}
To verify that \eqref{CFIDcov} quantifies the error in $\Sigxuhgyu$, notice that \eqref{CFIDcov} equals zero when $\Sigxuhgyu=\Sigxugyu$ and is otherwise positive (by Cauchy Schwarz).

\begin{table}[t]
  \caption{The mean and covariance components of CFID, along with the total CFID, for the generative models in the MRI and inpainting experiments. 
  For the MRI experiment, CFID$^1$ used 72 test samples and $P=32$,
  CFID$^2$ used 2\,376 test samples and $P=8$, and
  CFID$^3$ used all 14\,576 samples and $P=1$.
  For the inpainting experiment, CFID$^1$ used $1\,000$ test images and $P=32$,
  CFID$^2$ used $3\,000$ test and validation images and $P=8$, and
  CFID$^3$ used all 30\,000 images and $P=1$.}
  \vspace{0.1in}
  \label{tab:cfiderrors}
  \centering
  \resizebox{\columnwidth}{!}{%
  \begin{tabular}{llllllllll}
    \toprule
    Model 
        & CFID$^1\mean\!\downarrow$ & CFID$^1\cova\!\downarrow$ & CFID$^1\!\downarrow$ 
        & CFID$^2\mean\!\downarrow$ & CFID$^2\cova\!\downarrow$ & CFID$^2\!\downarrow$
        & CFID$^3\mean\!\downarrow$ & CFID$^3\cova\!\downarrow$ & CFID$^3\!\downarrow$\\
    \midrule
    & \multicolumn{9}{c}{$R=4$ MRI} \\ \cmidrule(r){2-10}
    Langevin (Jalal \cite{Jalal:NIPS:21})    & 1.89 & 3.40 & 5.29 & - & - & - & - & - & -\\
    cGAN (Adler \cite{Adler:18})         & 3.12 & 3.27 & 6.39 & 2.79 & 1.48 & 4.27 & 2.71 & 1.10 & 3.82\\
    cGAN (Ohayon \cite{Ohayon:ICCVW:21}) & 1.94 & 2.12 & 4.06 & 2.27 & 1.00 & 3.27 & 2.29 & 0.66 & 2.95\\
    cGAN (Ours) & \bf 0.98 & \bf 2.12 & \bf 3.10 & \bf 0.86 & \bf 0.68 & \bf 1.54 & \bf 0.86 & \bf 0.43 & \bf 1.29\\
    & \multicolumn{9}{c}{$R=8$ MRI} \\ \cmidrule(r){2-10}
    Langevin (Jalal \cite{Jalal:NIPS:21})    & 2.61 & 4.73 & 7.34 & - & - & - & - & - & -\\
    cGAN (Adler \cite{Adler:18})         & 5.00 & 5.10 & 10.10 & 4.16 & 2.14 & 6.30 & 4.09 & 1.63 & 5.72\\
    cGAN (Ohayon \cite{Ohayon:ICCVW:21}) & 2.73 & 3.31 & 6.04 & 3.07 & 1.52 & 4.59 & 3.30 & 0.97 & 4.27\\
    cGAN (Ours) & \bf 1.55 & \bf 3.32 & \bf 4.87 & \bf 1.24 & \bf 0.99 & \bf 2.23 & \bf 1.17 & \bf 0.62 & \bf 1.79\\
    & \multicolumn{9}{c}{Inpainting} \\ \cmidrule(r){2-10}
    Score SDE (Song \cite{Song:ICLR:21}) & 0.97 & \bf 38.69 & \bf 39.66 & -    & -     & - & 0.90 & \bf 4.21 & 5.11\\
    CoModGAN (Zhao \cite{Zhao:ICLR:21})    & 0.42 & 41.21     & 41.63     & 0.35 & 25.39 & 25.74 & 0.32 & 4.98 & 5.29\\
    cGAN (Ours) & \bf 0.32 & 39.41 & 39.73 & \bf 0.25 & \bf 22.32 & \bf 22.58 & \bf 0.24 & 4.45 & \bf 4.69\\
    \bottomrule
  \end{tabular}
  }
\end{table}

In \tabref{cfiderrors}, we report $\CFID\mean$ and $\CFID\cova$ for the MRI and inpainting experiments, in addition to the total CFID (also shown in Tables~\ref{tab:mriresults} and \ref{tab:inpainting256}).
As before, we computed CFID on three test sets for each experiment, 
which contained 72, 2\,376, and 14\,576 samples respectively for MRI, and 
1000, 3000, and 30\,000 samples respectively for inpainting. 
Due to the slow sample-generation time of the Langevin/score-based methods \cite{Jalal:NIPS:21,Song:ICLR:21}, we did not have the computational resources to evaluate them on all datasets, and that's why certain table entries are blank.

For both MRI experiments, \tabref{cfiderrors} shows our method outperforming the competing methods in both the mean and covariance components of CFID (and thus the total CFID) for all sample sizes.
And, in the inpainting experiment, \tabref{cfiderrors} shows our method outperforming CoModGAN in both the mean and covariance components (and thus the total CFID) for all sample sizes.

For the inpainting experiment, \tabref{cfiderrors} shows our method outperforming the score-based approach in total CFID on the 3000- and 30\,000-sample test sets but not on the 1000-sample test set.
However, we now argue that the 1000-sample inpainting experiment is heavily affected by small-sample bias, and therefore untrustworthy.
Looking at the mean component of CFID (i.e., $\CFID\mean^1$, $\CFID\mean^2$, and $\CFID\mean^3$) across the inpainting experiments, we see that the values are relatively small and stable with sample size.
But looking at the covariance component of CFID (i.e., $\CFID\cova^1$, $\CFID\cova^2$, and $\CFID\cova^3$) across the inpainting experiments, we see that the values are large and heavily dependent on sample size.
For the 1000-sample inpainting experiment, the total CFID is dominated by the covariance component and thus strongly affected by small-sample bias.
Consequently, for the 1000-sample inpainting experiment, the total CFID is not trustworthy.



\clearpage

\newcommand{\MRIzoomedappfig}[6]{
    \begin{figure}[h]
    \def\tabwid{18mm} 
    \def\figwid{\columnwidth}
    \rotatebox{0}{\sf \scriptsize \hspace{1.25mm}
        \begin{tabular}{>{\centering}p{\tabwid}>{\centering}p{\tabwid}>{\centering}p{\tabwid}>{\centering}p{\tabwid}>{\centering}p{\tabwid}>{\centering}p{\tabwid}}
         & E2E-VarNet & cGAN (ours) & cGAN (Ohayon) & cGAN (Adler) & Langevin (Jalal) 
        \end{tabular}}\\[-0.75mm]
    \rotatebox{90}{\sf \scriptsize
        \begin{tabular}{>{\centering}p{\tabwid}>{\centering}p{\tabwid}>{\centering}p{\tabwid}>{\centering}p{\tabwid}>{\centering}p{\tabwid}>{\centering}p{\tabwid}}
        & Truth & Truth &  &  &  
        \end{tabular}}
    \hspace{-2mm}
    \begin{tikzpicture}
        \node[anchor=south west,inner sep=0] at (0,0) {\includegraphics[width=0.95\columnwidth,trim=10 10 7 7,clip]{figures/mri/app_figs/app_mri_R_#6_fig_#1.png}};
        \draw [-latex, line width=1pt, yellow] (\fpeval{#2 - 4.4},\fpeval{#3 + 4.4}) -- (\fpeval{#4 - 4.4},\fpeval{#5 + 4.4});
        
        \draw [-latex, line width=1pt, yellow] (\fpeval{#2 - 2.2},\fpeval{#3 + 8.8}) -- (\fpeval{#4 - 2.2},\fpeval{#5 + 8.8});
        
        \draw [-latex, line width=1pt, yellow] (#2,\fpeval{#3 + 8.8}) -- (#4,\fpeval{#5 + 8.8});
        \draw [-latex, line width=1pt, yellow] (#2,\fpeval{#3 + 6.6}) -- (#4,\fpeval{#5 + 6.6});
        \draw [-latex, line width=1pt, yellow] (#2,\fpeval{#3 + 4.4}) -- (#4,\fpeval{#5 + 4.4});
        \draw [-latex, line width=1pt, yellow] (#2,\fpeval{#3 + 2.2}) -- (#4,\fpeval{#5 + 2.2});
        \draw [-latex, line width=1pt, yellow] (#2,#3) -- (#4,#5);

        \draw [-latex, line width=1pt, yellow] (\fpeval{#2 + 2.2},\fpeval{#3 + 8.8}) -- (\fpeval{#4 + 2.2},\fpeval{#5 + 8.8});
        \draw [-latex, line width=1pt, yellow] (\fpeval{#2 + 2.2},\fpeval{#3 + 6.6}) -- (\fpeval{#4 + 2.2},\fpeval{#5 + 6.6});
        \draw [-latex, line width=1pt, yellow] (\fpeval{#2 + 2.2},\fpeval{#3 + 4.4}) -- (\fpeval{#4 + 2.2},\fpeval{#5 + 4.4});
        \draw [-latex, line width=1pt, yellow] (\fpeval{#2 + 2.2},\fpeval{#3 + 2.2}) -- (\fpeval{#4 + 2.2},\fpeval{#5 + 2.2});
        \draw [-latex, line width=1pt, yellow] (\fpeval{#2 + 2.2},#3) -- (\fpeval{#4 + 2.2},#5);

        \draw [-latex, line width=1pt, yellow] (\fpeval{#2 + 4.4},\fpeval{#3 + 8.8}) -- (\fpeval{#4 + 4.4},\fpeval{#5 + 8.8});
        \draw [-latex, line width=1pt, yellow] (\fpeval{#2 + 4.4},\fpeval{#3 + 6.6}) -- (\fpeval{#4 + 4.4},\fpeval{#5 + 6.6});
        \draw [-latex, line width=1pt, yellow] (\fpeval{#2 + 4.4},\fpeval{#3 + 4.4}) -- (\fpeval{#4 + 4.4},\fpeval{#5 + 4.4});
        \draw [-latex, line width=1pt, yellow] (\fpeval{#2 + 4.4},\fpeval{#3 + 2.2}) -- (\fpeval{#4 + 4.4},\fpeval{#5 + 2.2});
        \draw [-latex, line width=1pt, yellow] (\fpeval{#2 + 4.4},#3) -- (\fpeval{#4 + 4.4},#5);

        \draw [-latex, line width=1pt, yellow] (\fpeval{#2 + 6.6},\fpeval{#3 + 8.8}) -- (\fpeval{#4 + 6.6},\fpeval{#5 + 8.8});
        \draw [-latex, line width=1pt, yellow] (\fpeval{#2 + 6.6},\fpeval{#3 + 6.6}) -- (\fpeval{#4 + 6.6},\fpeval{#5 + 6.6});
        \draw [-latex, line width=1pt, yellow] (\fpeval{#2 + 6.6},\fpeval{#3 + 4.4}) -- (\fpeval{#4 + 6.6},\fpeval{#5 + 4.4});
        \draw [-latex, line width=1pt, yellow] (\fpeval{#2 + 6.6},\fpeval{#3 + 2.2}) -- (\fpeval{#4 + 6.6},\fpeval{#5 + 2.2});
        \draw [-latex, line width=1pt, yellow] (\fpeval{#2 + 6.6},#3) -- (\fpeval{#4 + 6.6},#5);
        
    \end{tikzpicture}
    \rotatebox{90}{\sf \scriptsize
        \begin{tabular}{>{\centering}p{\tabwid}>{\centering}p{\tabwid}>{\centering}p{\tabwid}>{\centering}p{\tabwid}>{\centering}p{\tabwid}>{\centering}p{\tabwid}}
        Sample & Sample & Average (P=2) & Average (P=4) & Average (P=32) & Std. Dev. Map 
        \end{tabular}}
    \caption{Example $R=#6$ MRI reconstruction.
    Row one: pixel-wise SD with $P=32$, 
    Row two: $\hvec{x}\avgP$ with $P=32$,
    Row three: $\hvec{x}\avgP$ with $P=4$,
    Row four: $\hvec{x}\avgP$ with $P=2$,
    Rows five and six: posterior samples.
    The arrows indicate regions of meaningful variation across posterior samples.} 
    \label{fig:mriapp_R#6_#1}
    \end{figure}
}

\newcommand{\MRIzoomedappfigextraarrows}[9]{
    \begin{figure}[h]
    \def\tabwid{18mm} 
    \def\figwid{\columnwidth}
    \rotatebox{0}{\sf \scriptsize \hspace{1.25mm}
        \begin{tabular}{>{\centering}p{\tabwid}>{\centering}p{\tabwid}>{\centering}p{\tabwid}>{\centering}p{\tabwid}>{\centering}p{\tabwid}>{\centering}p{\tabwid}}
         & E2E-VarNet & cGAN (ours) & cGAN (Ohayon) & cGAN (Adler) & Langevin (Jalal)
        \end{tabular}}\\[-0.75mm]
    \rotatebox{90}{\sf \scriptsize
        \begin{tabular}{>{\centering}p{\tabwid}>{\centering}p{\tabwid}>{\centering}p{\tabwid}>{\centering}p{\tabwid}>{\centering}p{\tabwid}>{\centering}p{\tabwid}}
        & Truth & Truth &  &  &  
        \end{tabular}}
    \hspace{-2mm}
    \begin{tikzpicture}
        \node[anchor=south west,inner sep=0] at (0,0) {\includegraphics[width=0.95\columnwidth,trim=10 10 7 7,clip]{figures/mri/app_figs/app_mri_fig_#1.png}};
        \draw [-latex, line width=1pt, yellow] (\fpeval{#2 - 4.4},\fpeval{#3 + 4.4}) -- (\fpeval{#4 - 4.4},\fpeval{#5 + 4.4});
        
        \draw [-latex, line width=1pt, yellow] (\fpeval{#2 - 2.2},\fpeval{#3 + 8.8}) -- (\fpeval{#4 - 2.2},\fpeval{#5 + 8.8});
        
        \draw [-latex, line width=1pt, yellow] (#2,\fpeval{#3 + 8.8}) -- (#4,\fpeval{#5 + 8.8});
        \draw [-latex, line width=1pt, yellow] (#2,\fpeval{#3 + 6.6}) -- (#4,\fpeval{#5 + 6.6});
        \draw [-latex, line width=1pt, yellow] (#2,\fpeval{#3 + 4.4}) -- (#4,\fpeval{#5 + 4.4});
        \draw [-latex, line width=1pt, yellow] (#2,\fpeval{#3 + 2.2}) -- (#4,\fpeval{#5 + 2.2});
        \draw [-latex, line width=1pt, yellow] (#2,#3) -- (#4,#5);

        \draw [-latex, line width=1pt, yellow] (\fpeval{#2 + 2.2},\fpeval{#3 + 8.8}) -- (\fpeval{#4 + 2.2},\fpeval{#5 + 8.8});
        \draw [-latex, line width=1pt, yellow] (\fpeval{#2 + 2.2},\fpeval{#3 + 6.6}) -- (\fpeval{#4 + 2.2},\fpeval{#5 + 6.6});
        \draw [-latex, line width=1pt, yellow] (\fpeval{#2 + 2.2},\fpeval{#3 + 4.4}) -- (\fpeval{#4 + 2.2},\fpeval{#5 + 4.4});
        \draw [-latex, line width=1pt, yellow] (\fpeval{#2 + 2.2},\fpeval{#3 + 2.2}) -- (\fpeval{#4 + 2.2},\fpeval{#5 + 2.2});
        \draw [-latex, line width=1pt, yellow] (\fpeval{#2 + 2.2},#3) -- (\fpeval{#4 + 2.2},#5);

        \draw [-latex, line width=1pt, yellow] (\fpeval{#2 + 4.4},\fpeval{#3 + 8.8}) -- (\fpeval{#4 + 4.4},\fpeval{#5 + 8.8});
        \draw [-latex, line width=1pt, yellow] (\fpeval{#2 + 4.4},\fpeval{#3 + 6.6}) -- (\fpeval{#4 + 4.4},\fpeval{#5 + 6.6});
        \draw [-latex, line width=1pt, yellow] (\fpeval{#2 + 4.4},\fpeval{#3 + 4.4}) -- (\fpeval{#4 + 4.4},\fpeval{#5 + 4.4});
        \draw [-latex, line width=1pt, yellow] (\fpeval{#2 + 4.4},\fpeval{#3 + 2.2}) -- (\fpeval{#4 + 4.4},\fpeval{#5 + 2.2});
        \draw [-latex, line width=1pt, yellow] (\fpeval{#2 + 4.4},#3) -- (\fpeval{#4 + 4.4},#5);

        \draw [-latex, line width=1pt, yellow] (\fpeval{#2 + 6.6},\fpeval{#3 + 8.8}) -- (\fpeval{#4 + 6.6},\fpeval{#5 + 8.8});
        \draw [-latex, line width=1pt, yellow] (\fpeval{#2 + 6.6},\fpeval{#3 + 6.6}) -- (\fpeval{#4 + 6.6},\fpeval{#5 + 6.6});
        \draw [-latex, line width=1pt, yellow] (\fpeval{#2 + 6.6},\fpeval{#3 + 4.4}) -- (\fpeval{#4 + 6.6},\fpeval{#5 + 4.4});
        \draw [-latex, line width=1pt, yellow] (\fpeval{#2 + 6.6},\fpeval{#3 + 2.2}) -- (\fpeval{#4 + 6.6},\fpeval{#5 + 2.2});
        \draw [-latex, line width=1pt, yellow] (\fpeval{#2 + 6.6},#3) -- (\fpeval{#4 + 6.6},#5);

        \draw [-latex, line width=1pt, red] (\fpeval{#6 + 6.6},\fpeval{#7 + 8.8}) -- (\fpeval{#8 + 6.6},\fpeval{#9 + 8.8});
        \draw [-latex, line width=1pt, red] (\fpeval{#6 + 6.6},\fpeval{#7 + 6.6}) -- (\fpeval{#8 + 6.6},\fpeval{#9 + 6.6});
        \draw [-latex, line width=1pt, red] (\fpeval{#6 + 6.6},\fpeval{#7 + 4.4}) -- (\fpeval{#8 + 6.6},\fpeval{#9 + 4.4});
        \draw [-latex, line width=1pt, red] (\fpeval{#6 + 6.6},\fpeval{#7 + 2.2}) -- (\fpeval{#8 + 6.6},\fpeval{#9 + 2.2});
        \draw [-latex, line width=1pt, red] (\fpeval{#6 + 6.6},#7) -- (\fpeval{#8 + 6.6},#9);
        
    \end{tikzpicture}
    \rotatebox{90}{\sf \scriptsize
        \begin{tabular}{>{\centering}p{\tabwid}>{\centering}p{\tabwid}>{\centering}p{\tabwid}>{\centering}p{\tabwid}>{\centering}p{\tabwid}>{\centering}p{\tabwid}}
        Sample & Sample & Average (P=2) & Average (P=4) & Average (P=32) & Std. Dev. Map 
        \end{tabular}}
    \caption{Example $R=8$ MRI reconstruction.
    Row one: pixel-wise SD with $P=32$, 
    Row two: $\hvec{x}\avgP$ with $P=32$,
    Row three: $\hvec{x}\avgP$ with $P=4$,
    Row four: $\hvec{x}\avgP$ with $P=2$,
    Rows five and six: posterior samples.
    The yellow arrows indicate regions of meaningful variation across posterior samples.
    The red arrows show visible artifacts in the Langevin recovery.} 
    \label{fig:mriapp_#1}
    \end{figure}
}

\section{Additional reconstruction plots} 

\subsection{\texorpdfstring{$R=4$}{R=4} MRI Reconstruction} \label{app:mriplotsR4}
\MRIzoomedappfig{3}{5.3}{1.8}{5.45}{1.3}{4}
\MRIzoomedappfig{2}{4.9}{0.2}{5.05}{0.7}{4}
\MRIzoomedappfig{7}{4.9}{2.0}{5.05}{1.5}{4}
\MRIzoomedappfig{11}{5.8}{0.45}{5.3}{0.3}{4}
\MRIzoomedappfig{9}{5.4}{0.35}{5.9}{0.5}{4}
\MRIzoomedappfig{22}{5.65}{2.05}{5.15}{1.9}{4}

\clearpage
\subsection{\texorpdfstring{$R=8$}{R=8} MRI Reconstruction} \label{app:mriplotsR8}
\MRIzoomedappfig{3}{5.3}{1.8}{5.45}{1.3}{8}
\MRIzoomedappfig{2}{4.9}{0.2}{5.05}{0.7}{8}
\MRIzoomedappfig{7}{4.9}{2.0}{5.05}{1.5}{8}
\MRIzoomedappfigextraarrows{11}{5.8}{0.45}{5.3}{0.3}{5.7}{1.5}{5.55}{1.0}
\MRIzoomedappfigextraarrows{9}{5.4}{0.35}{5.9}{0.5}{5.7}{1.95}{5.85}{1.45}
\MRIzoomedappfig{22}{5.65}{2.05}{5.15}{1.9}{8}

\clearpage
\subsection{Inpainting} \label{app:inpaintingplots}
\vspace{-4mm}

\inpaintingFigRow{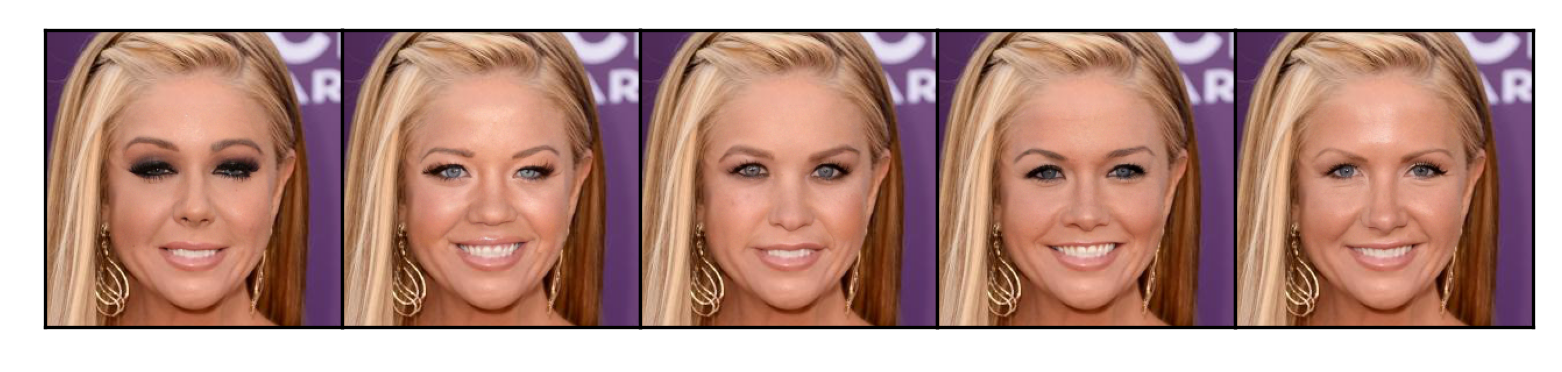}{h}
\inpaintingFigRow{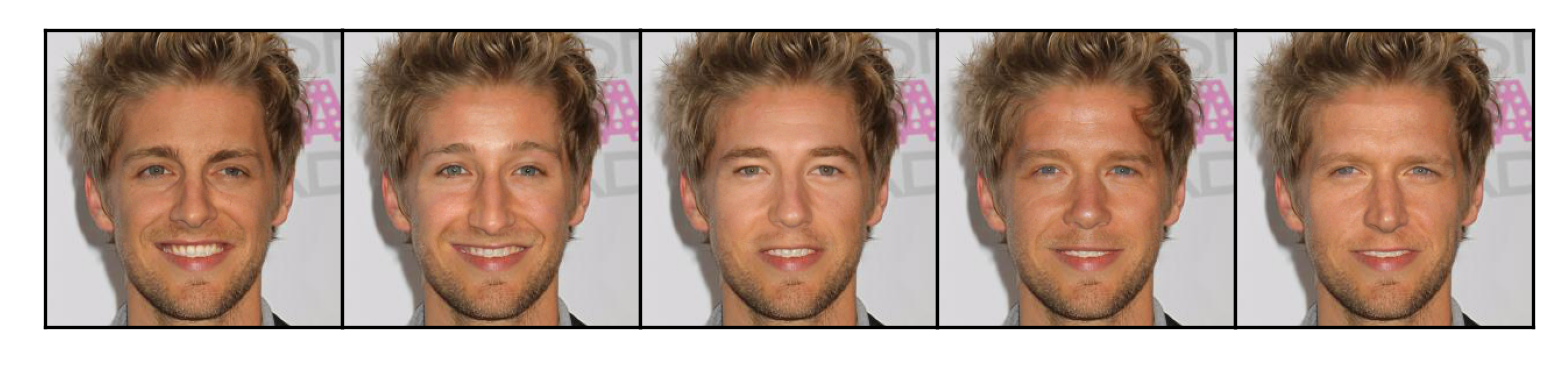}{h}
\inpaintingFigRow{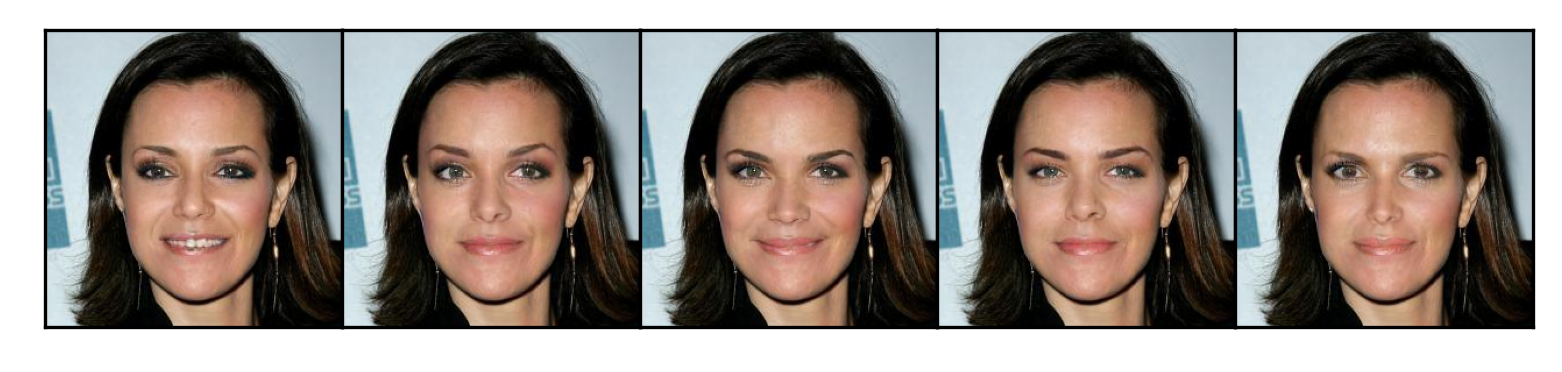}{h!}
\inpaintingFigRow{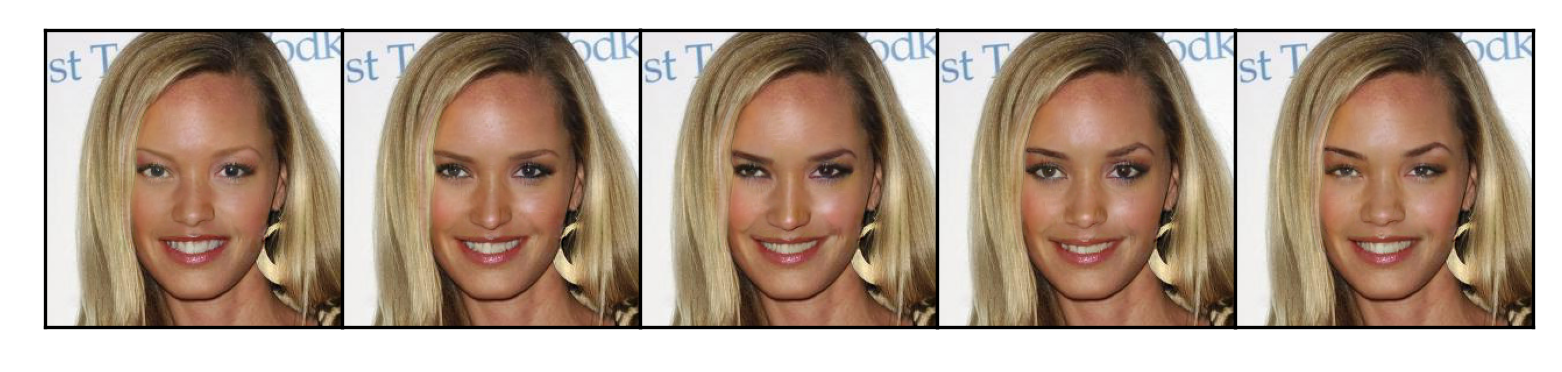}{t!}
\inpaintingFigRow{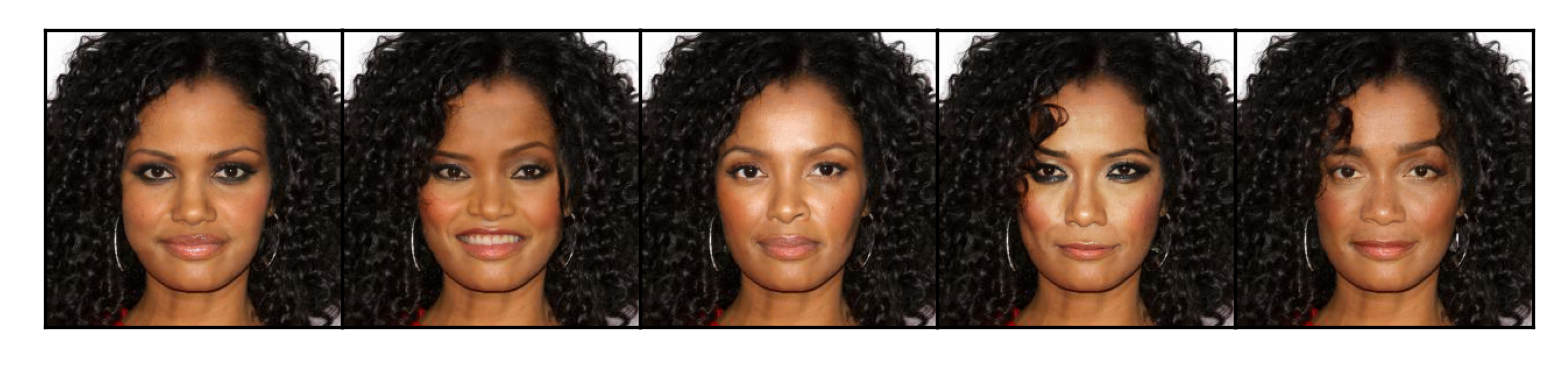}{h}
\inpaintingFigRow{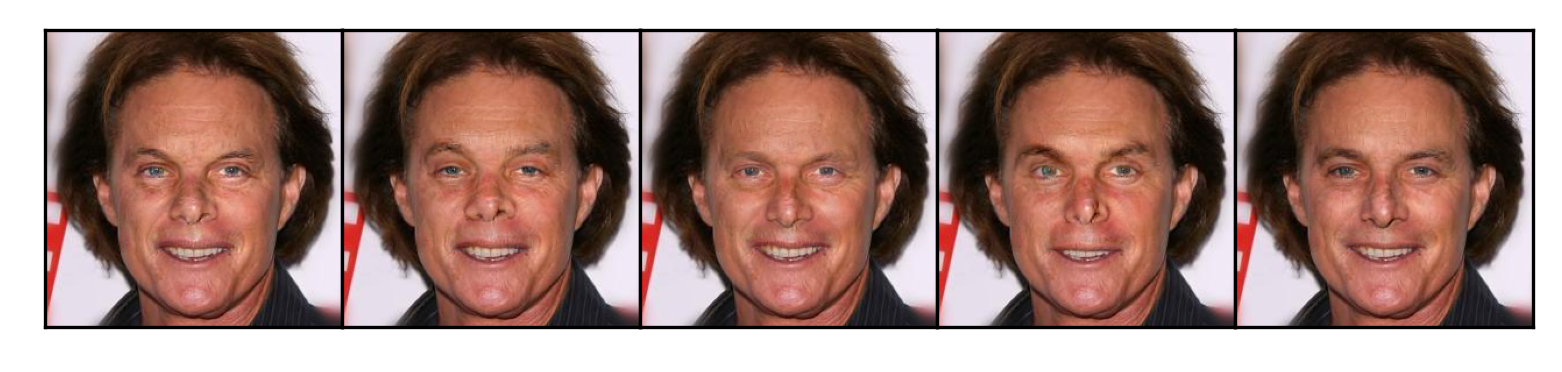}{h}

\end{document}